%% file: main.tex
\documentclass{article}

\PassOptionsToPackage{numbers, compress}{natbib}
\usepackage[preprint]{neurips_2024}

\usepackage[utf8]{inputenc} 
\usepackage[T1]{fontenc}    
\usepackage{hyperref}       
\usepackage{url}            
\usepackage{booktabs}       
\usepackage{amsfonts}       
\usepackage{nicefrac}       
\usepackage{microtype}      
\usepackage{xcolor}         
\usepackage{xspace}
\usepackage{amsmath}
\usepackage{amsthm}
\usepackage{enumitem}
\usepackage{xspace}
\usepackage[capitalize]{cleveref}
\usepackage{caption}
\usepackage{subcaption}
\usepackage{graphicx}
\usepackage{multirow}

\usepackage{algorithm}
\usepackage[noend]{algpseudocode}
\algnewcommand{\LineComment}[1]{\State \(\triangleright\) #1}

\hypersetup{
    colorlinks,
    linkcolor={blue!50!black},
    citecolor={blue!50!black},
    urlcolor={blue!50!black}
}
\renewcommand{\cite}{\citep}  

\include{macros}

\title{Fine-Tuning Large Language Models with \\ User-Level Differential Privacy}

\author{%
  Zachary Charles \\
  Google Research \\
  Seattle, WA, USA \\
  \texttt{zachcharles@google.com} \\
  \And  
  Arun Ganesh \\
  Google Research \\
  Seattle, WA, USA \\
  \texttt{arunganesh@google.com} \\
  \And
  Ryan McKenna \\
  Google Research \\
  Seattle, WA, USA \\
  \texttt{mckennar@google.com} \\
  \AND
  H. Brendan McMahan \\
  Google Research \\
  Seattle, WA, USA \\
  \texttt{mcmahan@google.com}
  \And
  Nicole Mitchell \\
  Google Research \\
  San Francisco, CA, USA \\
  \texttt{nicolemitchell@google.com} \\
  \AND
  Krishna Pillutla \\
  IIT Madras \\
  Chennai, India \\
  \texttt{krishnap@dsai.iitm.ac.in} \\
  \And
  Keith Rush \\
  Google Research \\
  Seattle, WA, USA \\
  \texttt{krush@google.com} \\
}

\begin{document}

\maketitle

\begin{abstract}
We investigate practical and scalable algorithms for training large language models (LLMs) with user-level differential privacy (DP) in order to provably safeguard all the examples contributed by each user. We study two variants of DP-SGD with: (1) example-level sampling (\els) and per-example gradient clipping, and (2) user-level sampling (\uls) and per-user gradient clipping.
We derive a novel user-level DP accountant that allows us to compute provably tight privacy guarantees for \els.
Using this, we show that while \els can outperform \uls in specific settings, \uls generally yields better results when each user has a diverse collection of examples.
We validate our findings through experiments in synthetic mean estimation and LLM fine-tuning tasks under fixed compute budgets. 
We find that \uls is significantly better in settings where either (1) strong privacy guarantees are required, or (2) the compute budget is large.
Notably, our focus on LLM-compatible training algorithms allows us to scale to models with hundreds of millions of parameters and datasets with hundreds of thousands of users.
\end{abstract}

\section{Introduction}

Fully realizing the promise of large language models (LLMs) in a variety of domains often requires fine-tuning them on domain-specific data~\cite{scao2021data,lester2021power,bhatia2023tart}. In settings such as AI agents~\cite{xi2023rise}, email assistants~\cite{chen2019gmail}, and smartphone keyboards~\cite{xu2023federated}, the respective in-domain data of agent-user interactions, emails, and texts are highly sensitive. Without appropriate safeguards, using such data comes with major privacy risks, especially since LLMs have been repeatedly shown to leak their training data~\cite{carlini2021extracting,carlini2023quantifying}.

Differential privacy (DP)~\cite{dwork2006calibrating} is a promising approach to mitigating these privacy concerns: it gives rigorous guarantees on privacy leakage and can effectively eliminate data leakage from LLMs~\cite{carlini2019secret}. DP is most commonly used to protect individual examples in the dataset. This is referred to as \emph{example-level} or \emph{item-level} DP. However, in the context of fine-tuning with user data, each user might contribute multiple correlated examples with similarities in language use, vocabulary, and contexts. Thus, example-level DP can fail to adequately protect a user's privacy from, for instance, membership inference attacks at the \emph{user-level}~\cite{song2019auditing,kandpal2023user}. However, DP can also be instantiated at the user level to safeguard all the data contributed by a user and provably mitigate such concerns. In this work, we are interested in scalable and practical approaches to fine-tuning LLMs with user-level DP. 

Many prior works on learning with user-level DP are theoretical~\cite{levy2021learning,bassily2023user,ghazi2023user,liu2024user}. 
These works aim for optimal convergence rates and rely on subroutines such as robust aggregation and outlier removal.
Scaling such algorithms to LLM training with high-dimensional models in clusters of accelerators remains challenging.
Meanwhile, prior empirical work on user-level DP has focused on federated learning~\cite{mcmahan2018learning,wei2021user,xu2023federated} for training small models on edge devices. Their design decisions are driven by the limited communication bandwidth and memory of smartphones~\cite{kairouz2021advances}. By contrast, efficiently training LLMs in a datacenter requires carefully batched operations to maximize accelerator throughput. Concurrently with our work, \citet{chua2024mindprivacyunituserlevel} studied the empirical performance of user-level DP methods for datacenter training. While their approach and findings are broadly similar to ours, we give a detailed discussion of the differences between these works in \cref{sec:related_work}.

A significant hurdle to the usage of DP at LLM-scale is its added computational cost~\citep{de2022unlocking}. There is a delicate trade-off in privacy, utility, and computational efficiency of DP algorithms. 
In this work, we explore these trade-offs at a range of fixed \emph{compute budgets}, which, in the context of LLMs, are often determined a priori based on system constraints such as accelerator availability.

\paragraph{Setting.}
We focus on two variants of DP-SGD that scale to LLM settings. In the first method, we apply DP-SGD~\citep{abadi2016deep} with example-level sampling and per-example gradient clipping. We refer to this as \elsfull, and abbreviate as \els. We apply \els to a dataset formed by pooling together at most $G_\elst$ examples from each user and translate this example-level DP guarantee to the user-level.
A similar user-to-example reduction was used in the theoretical work of~\citet{ghazi2023user}, although their algorithm runs in super-polynomial time in the worst case and is not scalable to LLMs.

In the second method, we apply DP-SGD with user-level sampling and per-user gradient clipping. We refer to this as \ulsfull, and abbreviate as \uls. In \uls, we compute gradients averaged over $G_\ulst$ examples from each sampled user, and apply DP-SGD to these ``user-level'' gradients. This approach underlies most prior theoretical work on user-level DP~\cite[e.g.][]{levy2021learning,bassily2023user,liu2024user} and the empirical work on federated learning (where it is known as DP-FedSGD~\citep{mcmahan2018learning}). Efficiently implementing \uls at LLM scales requires carefully sharding per-user gradient computations across user-partitioned data.

We focus on two primary questions in this work. First, both \els and \uls crucially rely on \emph{group size} parameters $G_\elst$ and $G_\ulst$, governing how many examples from each user are used within the algorithm's computations. We give theoretically and empirically justified heuristics for how to set these parameters in practical settings, for varying privacy and compute budgets. The second question is, given optimal (or near-optimal) configuration of these two methods, which one should a practitioner use when fine-tuning LLMs with user-level DP?

\paragraph{Contributions.} Our work provides both theoretical and empirical insights into scalable methods for training LLMs with user-level DP. First, we develop novel DP accounting techniques that allow us to provide user-level DP bounds for \elsfull. Our techniques are provably tight, and significantly outperform generic example-to-user reductions (e.g.~\citep{vadhan2017complexity}). Our key result is a tight analysis of the group privacy loss of iterated compositions of a sub-sampled Gaussian mechanism. We use these DP bounds to provide theoretical insight into the worst-case relative performance of \els and \uls.

Second, we provide detailed empirical comparisons of the two methods in fixed-compute settings. We use a synthetic mean estimation task to show that when configured correctly, \uls often outperforms \els, and the magnitude of improvement is largest when users have high variance across their gradients. We corroborate these findings in realistic fine-tuning tasks on a 350 million parameter transformer model. Throughout, we give best practices and empirically useful heuristics for configuring group sizes $G_\elst, G_\ulst$ for a compute budget. We demonstrate that by using LLM-tailored dataset pipelines and algorithm implementations, we can scale \els and \uls to models with hundreds of millions of parameters and datasets with hundreds of thousands of users. To the best of our knowledge, this constitutes the largest scale of empirical research on user-level DP to date.

\section{Algorithms and Optimal User-Level Privacy Accounting}

We first introduce the notion of user-level DP.
To define $(\varepsilon, \delta)$-DP, we use the $\alpha$-hockey-stick divergence $H_\alpha$ between distributions $P, Q$ and its symmetrized version $H_\alpha^{\sym}$:
\[
\textstyle
H_\alpha(P, Q) := \max_S \{P(S) - \alpha Q(S)\},
\quad\text{and}\quad H_\alpha^{\sym}(P, Q) = \max\{H_\alpha(P, Q), H_\alpha(Q, P)\}.
\]

Let $\calM$ be a randomized algorithm that takes in a dataset $D$ and returns a distribution $\calM(D)$. We say that $\calM$ satisfies $(\varepsilon, \delta)$-DP under a symmetric adjacency relation ``$\sim$'' if
\begin{equation}\label{eq:dp_definition}
\textstyle
    \sup_{D \sim D'} H^{\sym}_{e^\varepsilon}(\calM(D), \calM(D')) \leq \delta.
\end{equation}
\citet{barthe2013beyond} showed that this $(\epsilon, \delta)$-DP definition is equivalent to the more canonical definition of \citet{dwork2006our}. We use this DP definition because as noted by \cite{balle2018privacy}, it is more convenient for proofs such as those in Appendix~\ref{sec:accounting} that work closely with privacy loss distributions.

We now consider \emph{user-partitioned} datasets. Each dataset is a multiset, i.e., each datapoint is associated with a finite multiplicity.
Let $\calU$ be some index set of \emph{users}, such that each $u \in \calU$ is associated to some non-empty, finite dataset $D_u$. A user partitioned dataset is a tuple $(D, U)$ where $U \subseteq \calU$, $|U| < \infty$, and the full dataset $D = \sqcup_{u \in U} D_u$ is the multiset sum of the user datasets. 
We define $(D, U) \sim_{\text{user}} (D', U')$ if $U' = U \cup \{u\}$ or $U' = U \backslash \{u\}$ for some user $u$. We say that $\calM$ satisfies $(\varepsilon, \delta)$ user-level DP if $\calM$ satisfies \eqref{eq:dp_definition} w.r.t. the user-level adjacency relation $\sim_{\text{user}}$. When each user has a single example, this recovers the add-or-remove notion of example-level DP.

\paragraph{Algorithms for training with user-level DP.}
We describe two generalizations of DP-SGD to the setting of user-level DP. 
The first, DP-SGD with example-level sampling (\elsfull, abbreviated \els) first selects a subset of the data to train on.
Rather than training on the full dataset $D$, we train on a subset $D_{\text{sub}}$ to which each user contributes at most $G_\elst$ examples. We then perform DP-SGD on $D_{\text{sub}}$, with example-level sampling and gradient clipping. Below, we show how to translate the obtained example-level DP bound to a user-level DP bound.

The second algorithm, DP-SGD with user-level sampling (\ulsfull, abbreviated \uls), computes averaged gradients over $G_\ulst$ of each sampled user's examples. It clips these user gradients, rather than example gradients. The DP-FedSGD algorithm for federated learning~\citep{mcmahan2018learning} is a special case of \ulsfull.
\ulsfull can also be viewed as a variant of user-level DP algorithms proposed in \citep{levy2021learning, bassily2023user, liu2024user}, one that uses direct averaging to compute user-level gradients, rather than robust mean estimation.

We present \els and \uls in \Cref{alg:dp-sgd-els,alg:dp-sgd-uls}. They share common inputs: initial model $\theta^0$, loss function $f(\theta, z)$, user-level dataset $D = \sqcup_{u=1}^N D_u$, learning rate $\eta$, clip norm $C$, and number of iterations $T$. While we present an SGD model update in both, any first-order optimization technique can be applied, including momentum and learning rate adaptivity.

\begin{minipage}[t]{0.49\textwidth}
\begin{algorithm}[H]
\centering
\caption{\elsfull}\label{alg:dp-sgd-els}
\begin{algorithmic}
\State \textbf{Additional Inputs:}  
\State $\quad$ group size $G_\elst$, noise multiplier $\sigma_\elst$
\State $\quad$ example sampling probability $p$
\State $D_{\text{sub}} = \emptyset$ \Comment{\CommentStyle{Limit user contributions}}
\For{each user $u \in [N]$}
  \State Sample $S_u \subseteq D_u$ of size $|S_u| \le G_\elst$
  \State $D_{\text{sub}} = D_{\text{sub}} \sqcup S_u$
\EndFor
\State $B = p |D_\text{sub}|$ \Comment{\CommentStyle{Expected batch size}}
\For{$t = 0, 1, \ldots, T-1$}
  \LineComment{\CommentStyle{Include each \textbf{example} w.p. $p$}}
  \State Sample a batch of examples $S^t \subseteq D_{\text{sub}}$
  \LineComment{\CommentStyle{Clip \& noise \textbf{per-example} gradients}}
  \State $g^t_{\text{sum}} = \sum_{z \in S^t} \clip(\nabla f(\theta^t, z), C)$
  \State $g^t = \frac{1}{B} \left( g^t_{\text{sum}} + \calN(0, C^2 \sigma_\elst^2 I_d) \right)$
  
  \State $\theta^{t+1} = \theta^t - \eta g^t$
\EndFor
\end{algorithmic}
\end{algorithm}
\end{minipage}
\hfill
\begin{minipage}[t]{0.49\textwidth}
\begin{algorithm}[H]
\caption{\ulsfull}\label{alg:dp-sgd-uls}
\begin{algorithmic}
\State \textbf{Additional Inputs:}  
\State $\quad$ group size $G_\ulst$, noise multiplier $\sigma_\ulst$
\State $\quad$  user sampling probability $q$
\vspace{0.11em}
\State $M = q N$ \Comment{\CommentStyle{Expected user cohort size}}
\vspace{0.35em}
\For{$t = 0, 1, \ldots, T-1$}
  \LineComment{\CommentStyle{Include each \textbf{user} w.p. $q$}}
  \State Sample users $U^t \subseteq [N]$
  \vspace{0.25em}
  \For{each \textbf{user} $u \in U^t$}
    
    \State Sample $D_u^t \subseteq D_u$ of size $|D_u^t| \le G_\ulst$ 
    \State $g_u^t = \frac{1}{|D_u^t|} \sum_{z \in D_u^t} \nabla f(\theta^t, z)$
  \EndFor
  \LineComment{\CommentStyle{Clip \& noise \textbf{per-user} gradients}}
  \State $g^t_{\text{sum}} = \sum_{u \in U^t} \clip(g_u^t, C)$
  \State $g^t = \frac{1}{M} \left(g^t_{\text{sum}} + \calN(0, C^2 \sigma_\ulst^2  I_d)\right)$
  \State $\theta^{t+1} = \theta^t - \eta g^t$
\EndFor
\end{algorithmic}
\end{algorithm}
\end{minipage}

\paragraph{User-level DP accountants.}
The previous state-of-the-art accounting results for \cref{alg:dp-sgd-els} used black-box user-to-example reductions based on group privacy~\citep{vadhan2017complexity}. As a result, prior work~\cite[e.g.][]{levy2021learning}
suggests that \Cref{alg:dp-sgd-els} will be worse than \Cref{alg:dp-sgd-uls} as the formal $\varepsilon$ from group privacy grows quickly with $G_\elst$.
We instead tailor the reduction to DP-SGD. By leveraging the Mixture-of-Gaussians (MoG) mechanisms~\citep{choquette2023privacy}, we derive the optimal user-level DP accounting for \els.

\begin{thm}[Informal version of Theorem~\ref{thm:dp-sgd-accounting}]\label{thm:dp-sgd-accounting-informal}
For all $\varepsilon \geq 0$, Algorithm~\ref{alg:dp-sgd-els} satisfies $(\varepsilon, \delta(\varepsilon))$ user-level DP with
\begin{equation}\label{eq:DP-SGD-ELS-accounting}
\delta(\varepsilon) := H^{\sym}_{e^\varepsilon}(\calN(0, \sigma_\elst^2)^{\otimes T},~\calN(\bin(G_\elst, p), \sigma_\elst^2)^{\otimes T})\,,
\end{equation}
where $P^{\otimes T}$ denotes the product distribution $P \times \cdots \times P$ of any distribution $P$ with itself $T$ times.
For any $\delta' < \delta(\varepsilon)$, there is a setting where Algorithm~\ref{alg:dp-sgd-els} does not satisfy $(\varepsilon, \delta')$ user-level DP.
\end{thm}

\begin{figure}[htb]
    \centering
    \includegraphics[width=0.9\linewidth]{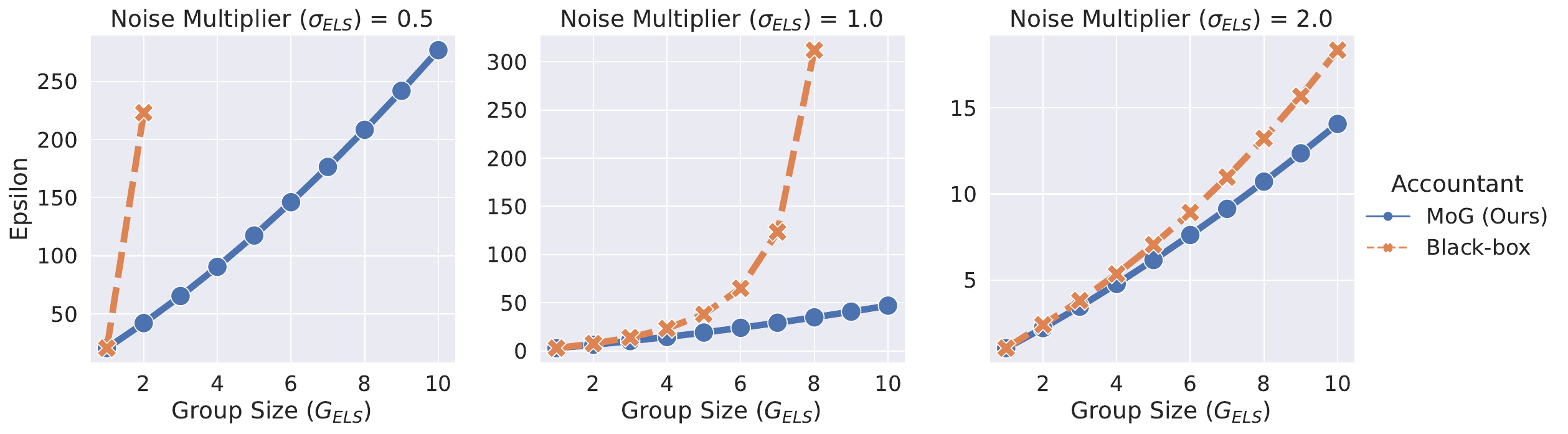}
    \caption{\small 
    Upper bound on $\epsilon$ for \els using our Mixture-of-Gaussians (MoG) accountant and the prior state-of-the-art black-box accounting from~\cite{vadhan2017complexity}. We set $T = 2000$ iterations, sampling probability $p = 10^{-2}$, $\delta = 10^{-6}$, and vary $\sigma_{\elst}$ and $G_{\elst}$. For $\sigma_{\elst} \in \{0.5, 1\}$, the black-box accountant diverged for $G_\elst$ sufficiently large.}
    \label{fig:els_accounting_comparison}
\end{figure}

The $\delta(\varepsilon)$ function in Theorem~\ref{thm:dp-sgd-accounting-informal} is easily computable using open-source DP accounting libraries (see Appendix~\ref{sec:accounting-implementation} for a code sample). 
We use this in \cref{fig:els_accounting_comparison} to empirically compare the $\epsilon$ computed by Theorem~\ref{thm:dp-sgd-accounting-informal} to the $\epsilon$ given by black-box reductions. \textbf{Our accountant gives near-linear scaling} of $\epsilon$ in $G_{\els}$ in settings whereas \textbf{generic user-to-example reductions give exponential scaling} in $G_{\els}$.
The optimality of our accountant is important for comparing \els to \uls: if \els is outperformed by \uls at some fixed user-level $(\varepsilon, \delta)$-DP guarantee, it is not because of slack in privacy accounting.

Since Algorithm \ref{alg:dp-sgd-uls} applies a straightforward subsampled Gaussian mechanism, its formal privacy guarantees are well understood and easy to compute \cite[e.g.][]{koskela21tight}. In order to directly compare \els and \uls, we will use the following result, which follows directly from Theorem \ref{thm:dp-sgd-accounting-informal} by setting $G_\elst = 1$.

\begin{thm}\label{thm:DP-SGD-ULS-accounting}
For all $\varepsilon \geq 0$, \Cref{alg:dp-sgd-uls} satisfies $(\varepsilon, \delta(\varepsilon))$ user-level DP with
\begin{equation}\label{eq:DP-SGD-ULS-accounting}
\delta(\varepsilon) := H^{\sym}_{e^\varepsilon}(\calN(0, \sigma_\ulst^2)^{\otimes T},~\calN(\bern(q), \sigma_\ulst^2)^{\otimes T}).
\end{equation}
For any $\delta' < \delta(\varepsilon)$, there is a setting where \Cref{alg:dp-sgd-uls} does not satisfy $(\varepsilon, \delta')$ user-level DP.
\end{thm}

\mypar{Comparisons under fixed compute budgets}  Both Algorithm \ref{alg:dp-sgd-els} and \ref{alg:dp-sgd-uls} are potentially reasonable choices for fine-tuning LLMs with user-level DP. However, the algorithms vary greatly in how they process data, and finding parameters that maximize utility for a given privacy-level may lead to spurious comparisons. Therefore, we focus on their relative performance under a \emph{fixed compute budget}. This is motivated by practical considerations: the massive size of LLMs often requires tailoring training pipelines to the compute budget, leading to the burgeoning field of scaling laws~\citep{kaplan2020scaling, hoffmann2022training}.

For simplicity, we assume the computation cost is equal to the total number of gradient computations. We do not consider the cost of gradient averaging, clipping, and noise generation, nor the cost of communication, though these can affect runtime.
The expected per-iteration compute of \els equals its expected batch size $B$, while that of \uls equals $G_\ulst M$, where $M= qN$ is the expected user cohort size (see \Cref{alg:dp-sgd-uls}).\footnote{For \uls, this is only an upper bound as some users may have fewer than $G_\ulst$ examples.}
Under a compute budget of $B$ gradients per iteration, we would thus aim for an expected user cohort size of $M = B / G_\ulst$. DP accounting of \els and \uls leverages amplification-by-sampling, but we use use shuffling and deterministic $B, M$ in our experiments\footnote{As discussed by ~\citet{ponomareva2023dpfy}, this practice introduces a possible gap between the reported bounds and the actual level of DP provided by the algorithm as implemented, though it is relatively common practice in the DP literature to date. Recent work of \citet{chua2024privatedpsgdimplementations} shows there is in fact an actual and unfavorable gap (shuffling can give worse privacy than Poisson sampling), and so while this practice remains reasonable for \emph{comparing} algorithms, it should not be used to report DP guarantees on actual privacy-sensitive datasets.}.

\section{Understanding \els and \uls with Lipschitz Losses} \label{sec:understanding}

We compare \els and \uls in a simplified setting with Lipschitz losses using their \emph{signal-to-noise ratio}, i.e., the noise variance per example gradient. This quantity is often heuristically used to tune DP-SGD hyperparameters (e.g. batch size), as it empirically correlates well with learning utility~\cite{ponomareva2023dpfy,sander2023tan}.

Consider Algorithms \ref{alg:dp-sgd-els} and \ref{alg:dp-sgd-uls}
operating on a user-partitioned dataset $D = \sqcup_{u=1}^N D_u$, where each user has $|D_u| = K$ examples.
To meet a compute budget of $T$ iterations with $B$ gradients per iteration (in expectation), we choose the sampling ratio $p=\nicefrac{B}{G_\elst N}$ for \els and $q = M/N = \nicefrac{B}{G_\ulst N}$ for \uls.
The practitioner is free to tune the group sizes $G_\elst, G_\ulst$. We analyze the impact of this choice.

Suppose the loss $f(\cdot, z)$ is Lipschitz for each $z$. Let $L_\elst, L_\ulst$ be the smallest constants such that 
\begin{equation}\label{eq:alpha_and_beta}
\max_{z\in D} \|\nabla f(\theta, z)\|_2 \le L_\elst, 
\qquad 
\max_{u \in [N]} \,\,  \max_{S \subset D_u : |S| = G_\ulst} \left\| \dfrac{1}{G_\ulst} \sum_{z \in S} \nabla f(\theta, z)\right\|_2 \le L_\ulst
\end{equation}
for all $\theta$.
Thus, $L_\elst$ and $L_\ulst$ respectively bound the norms of the per-example and per-user gradients.\footnote{
Note that $L_\elst$ and $L_\ulst$ are data-dependent. It is a common (albeit non-private) practice to tune the clip norm on the dataset. The problem of privately choosing a clip norm is beyond the scope of this work.}  We always have $L_\ulst \leq L_\elst$ by the triangle inequality. For simplicity of analysis, we set the clip norm $C$ to $L_\elst$ and  $L_\ulst$ for \els and \uls respectively, so the $\mathrm{clip}$ operation in \Cref{alg:dp-sgd-els,alg:dp-sgd-uls} is a no-op.
Then, the gradient estimates produced by \els and \uls at step $t$ are:
\begin{align}
g_{\elst}^t &= \dfrac{1}{B} \sum_{z \in S^t} \nabla f(\theta^t, z)  + \zeta_{\elst}^t,~~~\zeta_{\elst}^t \sim \mathcal{N}\left(0, \left(\frac{\sigma_{\elst}L_{\elst}}{B}\right)^2I_d\right)\,, \label{eq:els_gradient} \\
g_{\ulst}^t &= \dfrac{1}{B} \sum_{u\in U^t} \sum_{z \in D_u} \nabla f(\theta^t, z) + \zeta_{\ulst}^t,~~~\zeta_{\ulst}^t \sim \mathcal{N}\left(0, \left(\frac{\sigma_{\ulst}L_{\ulst}}{M}\right)^2I_d\right)\,. \label{eq:uls_gradient}
\end{align}

Here, the noise multipliers $\sigma_{\elst}$ and $\sigma_{\ulst}$ are determined by fixing a desired user-level DP guarantee $(\varepsilon, \delta)$ and a number of training steps $T$. Specifically, using Theorems~\ref{thm:dp-sgd-accounting-informal} and \ref{thm:DP-SGD-ULS-accounting} respectively, we set 
\begin{align*}
\sigma_{\elst} &= \inf\{\sigma: H^{\sym}_{e^\varepsilon}\big(\calN(0, \sigma^2)^{\otimes T},\calN(X, \sigma^2)^{\otimes T}\big) \leq \delta\} \qquad \text{for} \qquad X = \bin(G_\elst, p), \\
\sigma_{\ulst} &= \inf\{\sigma: H^{\sym}_{e^\varepsilon}\big(\calN(0, \sigma^2)^{\otimes T},\calN(Y, \sigma^2)^{\otimes T}\big) \leq \delta\} \qquad \text{for} \qquad Y = \bern(q).
\end{align*}
These noise multipliers are computable using open-source libraries, see Appendix~\ref{sec:accounting-implementation}.
If we compare the per-batch noise variances $\variance(\zeta_\elst^t)$ and $\variance(\zeta_\ulst^t)$ of updates \eqref{eq:els_gradient} and \eqref{eq:uls_gradient} respectively, we have, 
\begin{equation}\label{eq:sgd_beats_fedsgd}
    \variance(\zeta_\elst^t) \leq \variance(\zeta_\ulst^t) \iff L_\elst\sigma_\elst \leq G_\ulst L_\ulst\sigma_\ulst.
\end{equation}

While \eqref{eq:sgd_beats_fedsgd} is not entirely predictive of the relative performance of \els and \uls (since they sample data differently), 
it is a useful way to compare the methods.
\eat{
\begin{obs} \label{obs:uls_vs_els}
$G_{ULS} = 1$ implies $L_{ELS} = L_{ULS}$.  Moreover, it follows from \cref{lem:vector-to-scalar-reduction} that $\sigma_{ULS} \leq \sigma_{ELS}$ for all $\epsilon, \delta, T, G_{ELS}$.  Hence \uls with $G_{ULS}=1$ always enjoys a signal-to-noise ratio no larger than \els for any $G_{ELS}$.  Empirically, we have observed that $\sigma_{ELS} \approx \sigma_{ULS}$ in this setting and both methods thus have similar signal-to-noise ratios.
\end{obs}
}
The ratio of $L_\text{ELS}$ to $L_\text{ULS}$ is quite similar to the notion of \emph{gradient diversity}~\citep{yin2018gradient}. In general, \eqref{eq:sgd_beats_fedsgd} shows that when $L_\ulst$ is sufficiently small when compared to $L_\elst$, \uls adds noise with smaller variance than \els.

\begin{figure}[htb]
    \begin{subfigure}{0.3\linewidth}
    \centering
    \includegraphics[height=1.2in]{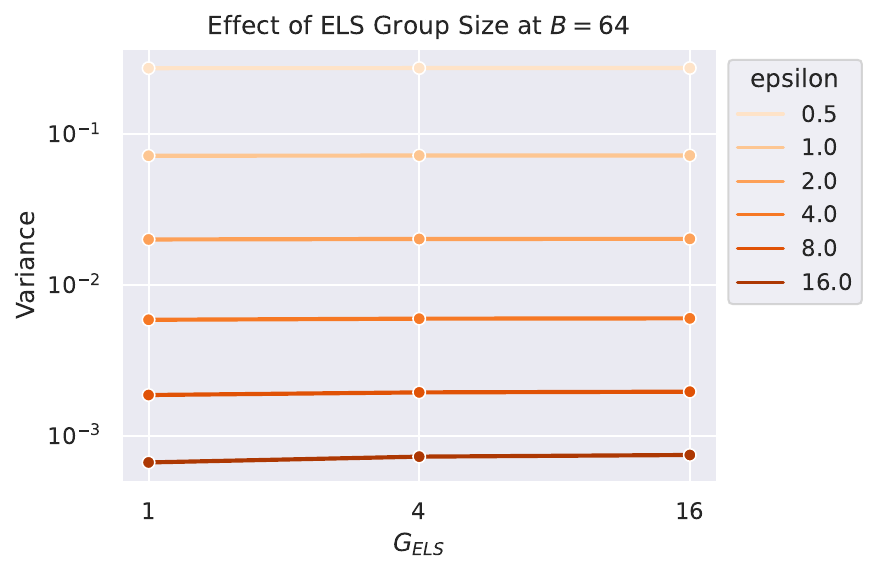}
    \caption{Noise variance of \els as a function of $G_\elst$.}
    \label{fig:els_group_size}
    \end{subfigure}
    \hspace{0.3cm}
    \begin{subfigure}{0.64\linewidth}
    \centering
    \includegraphics[height=1.2in]{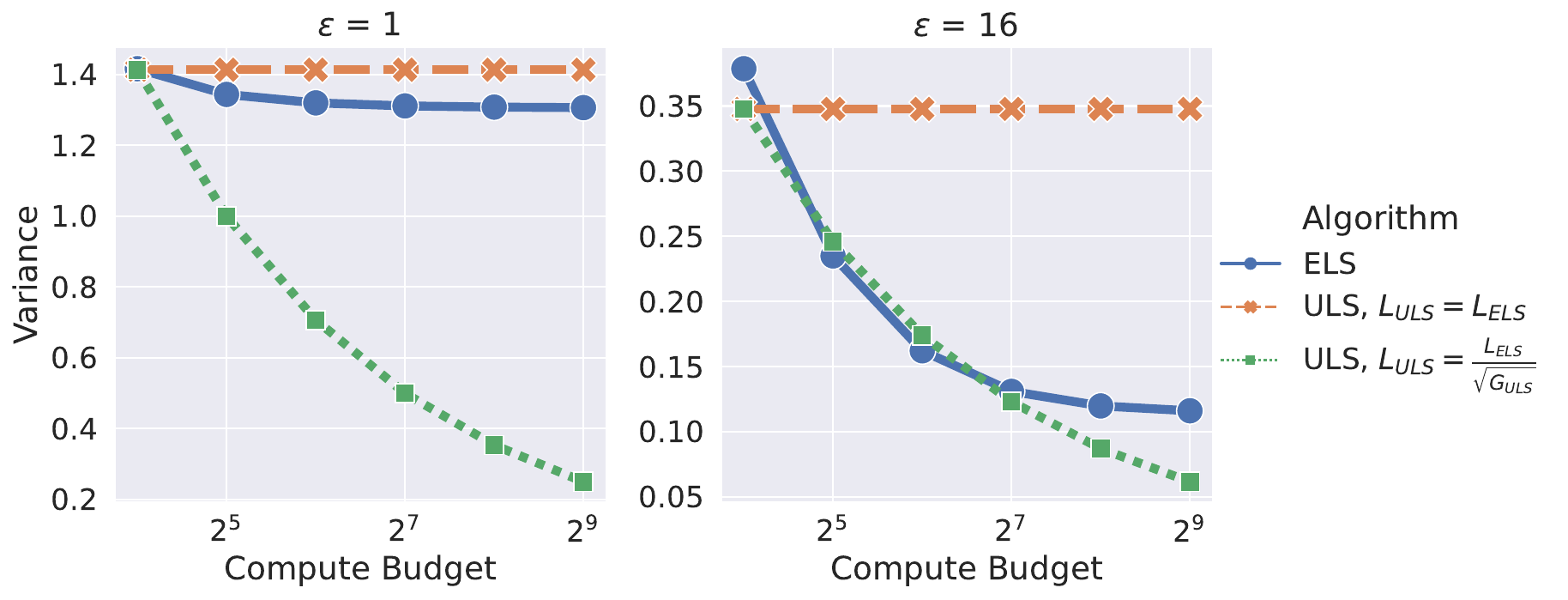}
    \caption{Noise variance of \els and \uls for varying compute budgets and different relative values of $L_\elst$ and $L_\ulst$.}
    \label{fig:uls_els_var}
    \end{subfigure}
    \caption{\small 
    A plot of the per-iteration variances $\variance(\zeta_{\elst})$ and $\variance(\zeta_{\ulst})$ of \els and \uls. We set $K = 32$, $T=1000$, and $N = 1024$. In (a) we fix a compute budget of $B = 64$ for $\els$ and vary $G_\elst$. In (b) For \els, we fix $G_\elst = K$ and the sampling probability $p = B / K N$ as we vary the compute budget $B$. For \uls, we fix the cohort size $M = 16$ and vary $G_\ulst = B / M$ with $B$. We compare two settings, one in which $L_\ulst = L_\elst$ (no intra-user gradient diversity), and one in which $L_\ulst = L_\elst / \sqrt{G_\ulst}$ (maximal intra-user diversity).}
    \label{fig:compare_variance}
\end{figure}

To illustrate this, we plot $\variance(\zeta_{\elst})$ and $\variance(\zeta_{\ulst})$ in \cref{fig:compare_variance}. \cref{fig:els_group_size} shows that $G_\elst$ has almost no effect on the variance $\variance(\zeta_{\elst})$. In \cref{fig:uls_els_var}, we see that \uls has smaller variance when $L_\ulst < L_\elst$ (specifically, $L_\ulst = L_\elst / \sqrt{G_\ulst}$, that is comparing the blue and green lines)  and either  $\varepsilon$ is small, or the compute budget $B$ is large (and so $G_\ulst$ is large) . \cref{appendix:compare_variances} contains details and more results.

\mypar{Comparison under maximal group sizes}
We now focus on the setting in which $G_\elst = G_\ulst = K$ where $K$ is the maximum user-dataset size. This represents the operating point at which per-example and per-user sampling are maximally different. By construction, in this setting $p = q = M/N$. We conjecture that when $L_{\elst} = L_{\ulst}$, $\variance(\zeta_{\elst}) \leq \variance(\zeta_{\ulst})$. This is worst-case for \uls, as it means that all gradients within a user point in the same direction. This conjecture is empirically supported by Figures \ref{fig:compare_variance} and \ref{fig:compare_variance_full}, by comparing variance at the largest compute budget for \els and \uls with $L_\elst = L_\ulst$.
\begin{conj}\label{conj:mog_accounting}
Let $G_\ulst = G_\elst = K$, and $p = q$. For all $\varepsilon, \delta, T, p$, and $K$, $\sigma_\elst \leq K\sigma_\ulst$.
\end{conj}
While this conjecture is challenging to prove for $(\varepsilon, \delta)$-DP, we show a weaker version of the conjecture holds for ``one-sided'' $\alpha$-RDP where $\alpha$ is an integer. Recall that for $\alpha > 1$, the $\alpha$-R\'enyi divergence between distributions $P, Q$ is defined by
$R_\alpha(P, Q) = (\nicefrac{1}{\alpha-1})\log\mathbb{E}_{x \sim Q}\left[\left(\nicefrac{P(x)}{Q(x)}\right)^\alpha\right]$.
\begin{lem}\label{lem:conj-ub}
Let $P_K(\sigma) = \calN(\bin(K, p), \sigma^2)$ and $Q(\sigma) = \calN(0, \sigma^2)$. For integers $\alpha > 1$, $K \geq 1$:
\[R_\alpha\big(P_K(K \sigma), Q(K \sigma)\big) \leq R_\alpha\big(P_1(\sigma), Q(\sigma)\big). \]
\end{lem}

To see how Lemma~\ref{lem:conj-ub} relates to \cref{conj:mog_accounting}, note that the RHS is approximately the R\'enyi-DP parameter of one iteration of \uls with noise multiplier $\sigma$ and the LHS is the R\'enyi-DP parameter of one iteration of \els with noise multiplier $K \sigma$. Since R\'enyi divergence is decreasing in the noise multiplier, this implies that there is some $\sigma' \leq K \sigma$ such that $R_\alpha(P_K(\sigma'), Q(\sigma')) = R_\alpha(P_1(\sigma), Q(\sigma))$, so \els requires a noise multiplier less than $K$ times that of \uls to obtain the same R\'enyi divergence bound. We give the proof in \cref{sec:conj-ub-proof}.
Intuitively, $\bern(p)$ only takes on its extreme values while $\bin(K, p) / K$ is more well-concentrated around $p$. Privacy is approximately a convex function of sensitivity, so the latter is better for privacy by Jensen's inequality.

\section{Synthetic Example: Mean Estimation}\label{sec:synthetic_task}

To better understand the behavior of \els and \uls, we evaluate them on a mean estimation task with a square distance loss. By focusing on this simple setting, we can thoroughly explore the factors that influence their relative performance, including dataset characteristics, compute budget, privacy budget, and algorithm hyperparameters. In contrast to \cref{sec:understanding}, the loss is not Lipschitz.

We first sample a population mean $\mu = \mathcal{N}(0, I_d)$, for $d = 32$. For each of $N = 256$ users, we sample a user mean $\mu_u = \mathcal{N}(\mu, \sigma_1^2 I_d)$, and user data $\{x_{u,j} \sim \mathcal{N}(\mu_u, \sigma_2^2 I_d)\}_{j=1}^{K}$ where $K = 16$. We refer to $\sigma_2$ as the ``within-user variance''. 
Our goal is to estimate $\mu$ under $(\epsilon, \delta)$-DP using \els or \uls.
To normalize compute, we set the cohort size $M$ in \uls as $M = \nicefrac{B}{G_\ulst}$. We fix $T = 256, \delta = 10^{-6}$ and $\sigma_1 = 1$. We use default values of $\epsilon = 1$, a per-iterate compute budget of $64$, and $\sigma_2 = 1$, but vary each separately to study how they affect the performance of \els and \uls. While we vary $G_\ulst$ in all experiments, we find use $G_\elst = K$ throughout (as it uniformly gives the best performance for \els). For each experiment setting, we sweep over learning rate and clip norm, and report the mean loss for the best setting across $128$ random trials.

\begin{figure}[htb]
    \centering
    \begin{subfigure}{0.32\linewidth}
        \includegraphics[width=\linewidth]{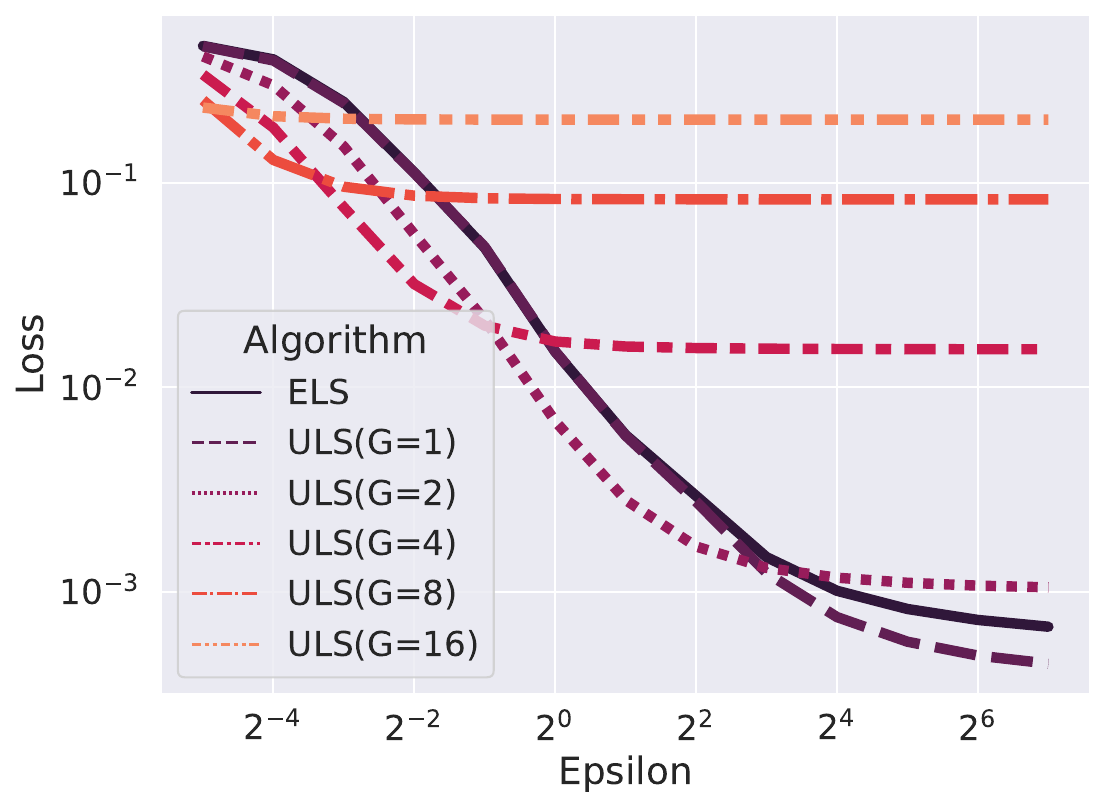}
        \caption{\small Varying $\varepsilon$.}
        \label{fig:synth_epsilon}
    \end{subfigure}
    \begin{subfigure}{0.32\linewidth}
        \includegraphics[width=\linewidth]{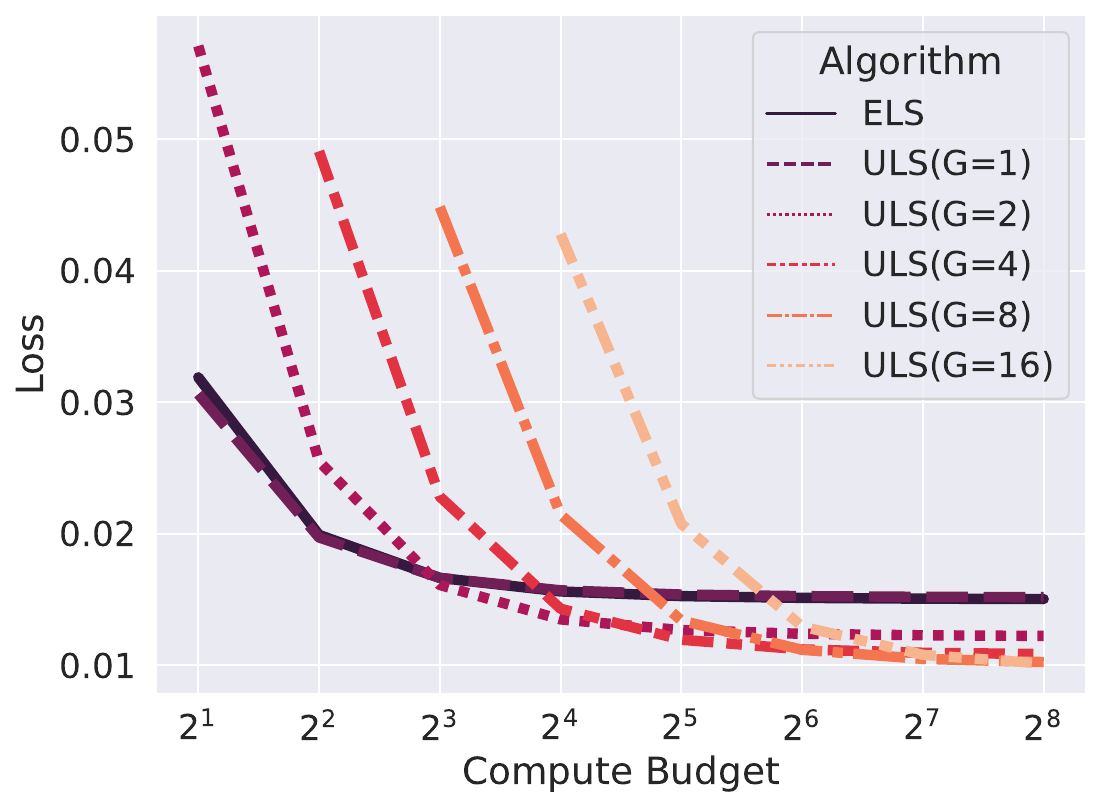}
        \caption{\small Varying compute budget.}
        \label{fig:synth_batch_size}
    \end{subfigure}
    \begin{subfigure}{0.32\linewidth}
        \includegraphics[width=\linewidth]{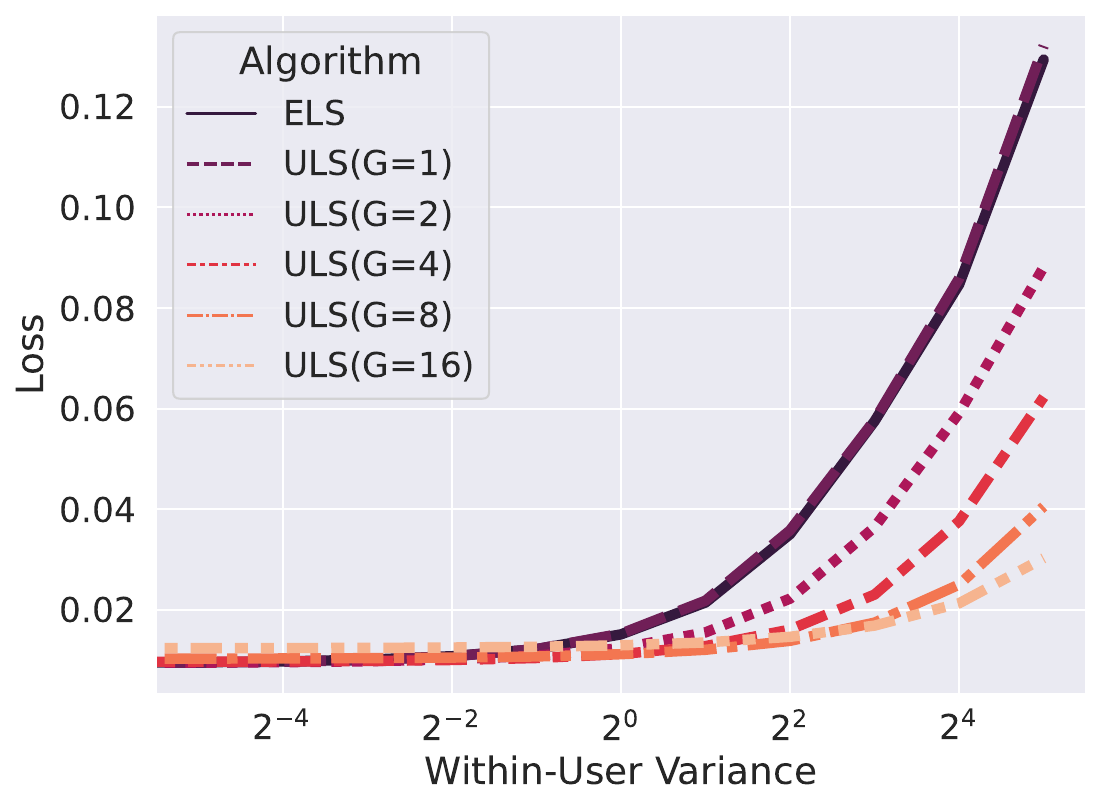}
        \caption{\small Varying within-user variance.}
        \label{fig:synth_wcv}
    \end{subfigure}
    \caption{\small Performance of \els and \uls on a synthetic mean estimation task with (a) varying $\varepsilon$, (b) varying compute budget, and (c) varying within-user variance $\sigma_2$. }
    \label{fig:synthetic}
\end{figure}

We visualize our results in \cref{fig:synthetic}. We find that \uls with $G_\ulst = 1$ is comparable to if not better than \els across all settings. 
We also see that \uls benefits from larger values of $G_\ulst$ when one of the following occurs: $\sigma_2$ (the within-user variance) is large, the compute budget (dictated by the batch size $B$) is large, or when $\epsilon$ is small.
In these regimes, \uls improves significantly on \els. We note that the notion that larger values of $G_\ulst$ can improve performance of \uls is corroborated theoretically for Lipschitz losses in \cref{sec:understanding}.

\section{Language Model Experimental Setup}\label{sec:exp-setup}

While \cref{sec:understanding} exhibits conditions under which \uls outperforms \els for user-level DP training, it is unclear whether this translates into benefits in realistic language model tasks. To investigate this, we apply both to language model fine-tuning, across a variety of privacy and compute budgets.

\mypar{Model}
We use a 350 million parameter decoder-only transformer model, with a WordPiece tokenizer, implemented in Praxis~\citep{praxis_lib}. We use a sequence length of 128, and train via a causal language modeling loss (i.e., next token prediction with cross-entropy loss).

\mypar{Datasets}
For fine-tuning, we use the Stack Overflow and CC-News datasets. Stack Overflow consists of questions and answers from the eponymous social media site, and is directly partitioned into users~\citep{stackoverflow}. The train split has 135,818,730 examples partitioned across 342,477 users. CC-News consists of English language articles on the web. We partition it according to the base domain of each article's URL. This results in 708,241 examples partitioned across 8,759 users.

For pre-training, we use a modified version of the C4 dataset~\citep{raffel2020exploring}. Because LLMs can memorize training data~\citep{carlini2021extracting}, we attempt to minimize
overlap between C4 and the fine-tuning datasets in order to more accurately reflect the advantages of DP finetuning. As noted by \citet{kurakin2023harnessing}, overlap between pre-training and fine-tuning may cause us to under-estimate how much DP fine-tuning can improve downstream performance. To remove overlap, we use a two-step filtering scheme: First, we use the NearDup method~\citep{lee2022deduplicating} to identify and remove approximate duplicates between C4 and the fine-tuning datasets. Second, we filter the URLs in what remains, removing all examples associated to \texttt{stackoverflow.com} or URLs contained in CC-News. We refer to the result as \cmm. For details, see \cref{appendix:datasets}.

\mypar{Training} We perform non-private pre-training of our transformer model on the \cmm dataset. We train for 400,000 steps, using a batch size of 512. We use the Adafactor optimizer~\cite{shazeer2018adafactor} with a cosine learning rate decay schedule. We tune the learning rate tuned based on C4 validation set performance. For fine-tuning, we use a dataset-dependent number of steps: 10,000 for Stack Overflow, and 2000 for CC-News. For both \els and \uls, we use Adafactor with a constant learning rate to perform model updates. We tune the learning rate and clip norm throughout. To decouple the two, we use the normalized clipping scheme proposed by \citet{de2022unlocking}.

\mypar{DP accounting and sampling} We vary $\varepsilon$ and set $\delta = n^{-1.1}$, where $n$ is the number of examples in the dataset (before any subsampling, as in \els), following \citep[Sec. 5.3.2]{ponomareva2023dpfy}. Thus, for Stack Overflow we set $\delta = 1.13\times 10^{-9}$, and for CC-News\footnote{For CC-News accounting, we make the assumption that there are $10\times$ more users than are actually present in the dataset, due to the smaller number of users. We revisit this choice later on.} we set $\delta = 2.92\times 10^{-8}$. We compute the noise multiplier assuming amplification via sub-sampling (at the example level for \els, and at the user level for \uls). For efficiency reasons, in practice we sample by shuffling. For Algorithm \ref{alg:dp-sgd-els}, we shuffle the dataset $D_{\text{sub}}$ and sample batches of a fixed size $B$ in shuffled order. For Algorithm \ref{alg:dp-sgd-uls}, we shuffle the set of users and sample user cohorts of a fixed size $M$ in shuffled order.

\mypar{Software and compute resources}
We define our model in Praxis~\citep{praxis_lib}. For \els, we use \texttt{tf.data} pipelines to create and iterate over datasets, and implement Algorithm \ref{alg:dp-sgd-els} in JAX. For \uls, we use Dataset Grouper~\citep{charles2023towards} to create and efficiently iterate over users in our datasets. \uls is implemented as a parallelized compute process in FAX~\citep{rush2024fax}, enabling linear scaling with respect to compute resources. All experiments were run in PAX~\citep{pax_lib}. We used TPU v3 pod slices, in $4\times 4$, $8\times 8$, and $16\times 16$ topologies for our small, medium, and large compute budget experiments.

\section{Language Model Results}\label{sec:exp-results}

\mypar{Tuning DP-SGD-ELS} In Algorithm \ref{alg:dp-sgd-els}, the group size $G_\elst$ involves a fundamental trade-off: If $G_\elst$ is small, then the resulting training dataset will be smaller and less diverse, as many clients may have more than $G_\elst$ examples. However, if $G_\elst$ is large, then we have to increase the noise multiplier to protect the privacy of users who contribute that many examples, even if most users have fewer than $G_\elst$ examples. Intuitively, the optimal setting of $G_\elst$ (in the sense of validation performance subject to a fixed privacy budget) would balance these two competing factors.

\begin{figure}[htb]
    \centering
    \includegraphics[width=0.46\linewidth]{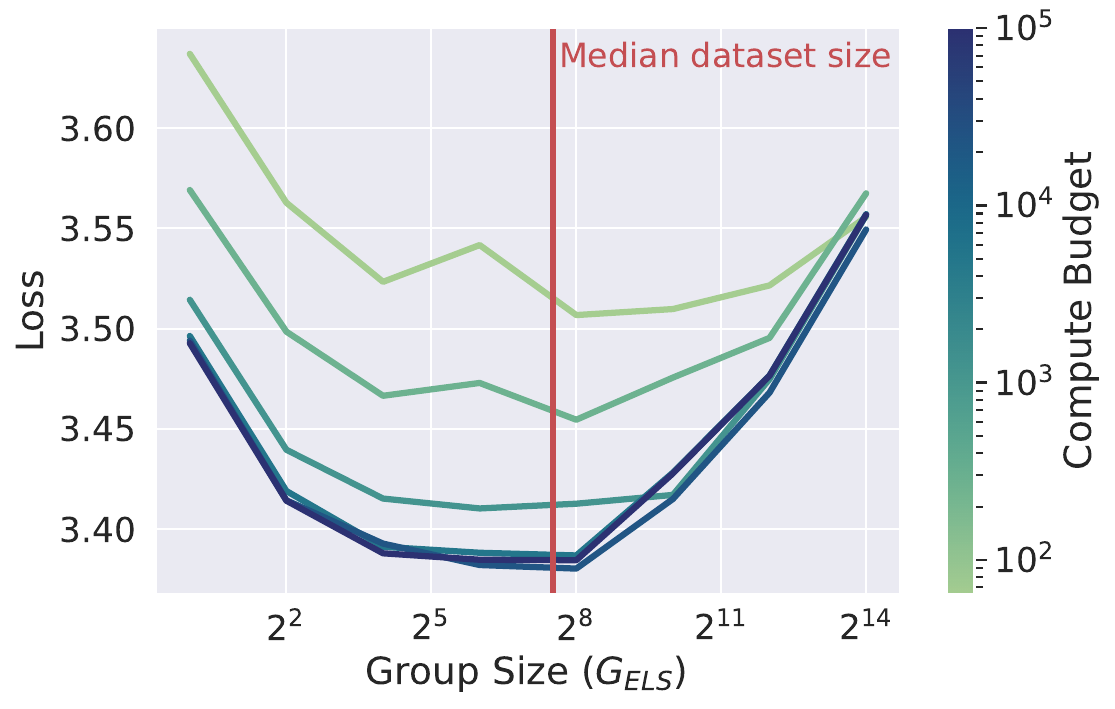}
    \includegraphics[width=0.46\linewidth]{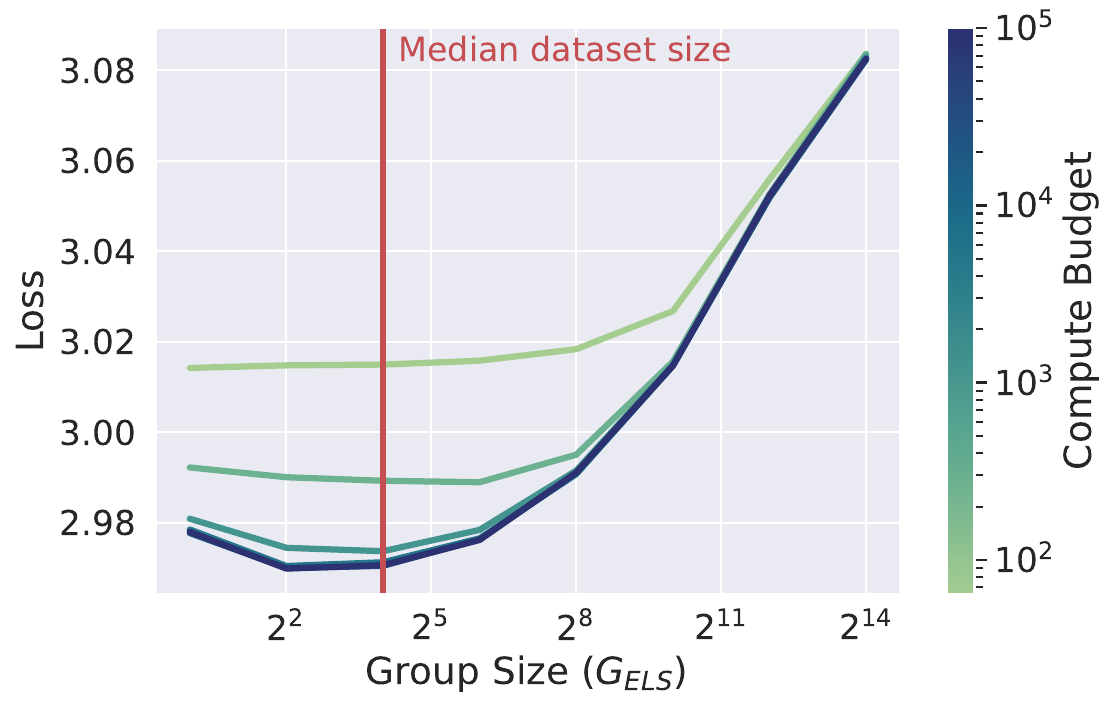}
    \caption{\small Validation loss of \els on Stack Overflow (left) and CC-News (right) for $\varepsilon = 4$ with varying compute budget and group size. The median user dataset size is plotted vertically in red.}
    \label{fig:els-group-size}
\end{figure}

To determine how to tune $G_\elst$, we fine-tune via \els, varying the compute budget and group size $G_\elst$, for a variety of $\epsilon$. The results for $\epsilon = 4$ are in Figure \ref{fig:els-group-size}. We observe a dataset-dependent range of group sizes that work well across compute budgets and $\varepsilon$. Setting $G_\elst$ to $2^7$ and $2^4$ works well for Stack Overflow and CC-News, respectively. This coincides almost exactly with the \emph{median} dataset size across all users.
As we show in \cref{appendix:configure_els}, similar group sizes work across values of $\varepsilon$. Throughout the remainder, we use this median heuristic to set $G_\elst$. This quantity can be approximated well using private median algorithms (see \cite{tzamos2020optimal}, for example).

\mypar{Tuning DP-SGD-ULS} Tuning DP-SGD-ULS requires tuning the group size $G_\ulst$. Recall that for a compute budget $B$, $G_\ulst M = B$ where $M$ is the cohort size. It is a priori unclear how to set these factors for a given compute budget. To understand this, we fine-tune while varying the cohort size $M$, group size $G_\ulst$, and privacy level $\varepsilon$. In all experiments, we have some compute budget $B = G_\ulst M$, and want to figure out how to allocate the compute budget across $G_\ulst$ and $M$ so as to minimize validation loss after fine-tuning with privacy budget $\epsilon$.

\begin{figure}[ht]
    \centering
    \begin{subfigure}[b]{0.4\textwidth}
        \centering
        \includegraphics[width=\textwidth]{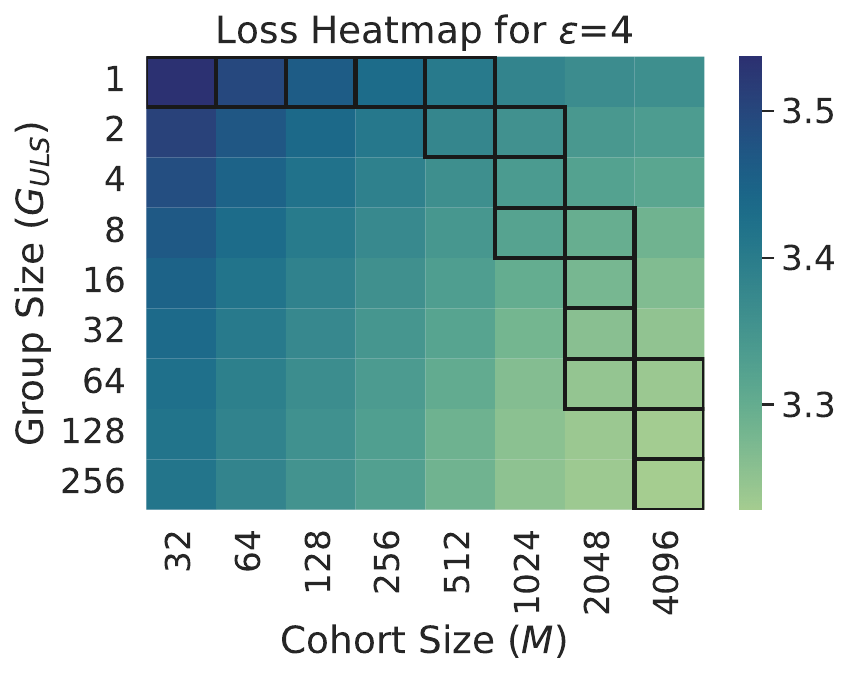}
        \caption{Loss heatmap. Optimal $G_\ulst, M$ for each compute budget are highlighted.}
        \label{fig:heatmap}
    \end{subfigure}
    \hfill
    \begin{subfigure}[b]{0.57\textwidth}
        \centering
        \includegraphics[width=\textwidth]{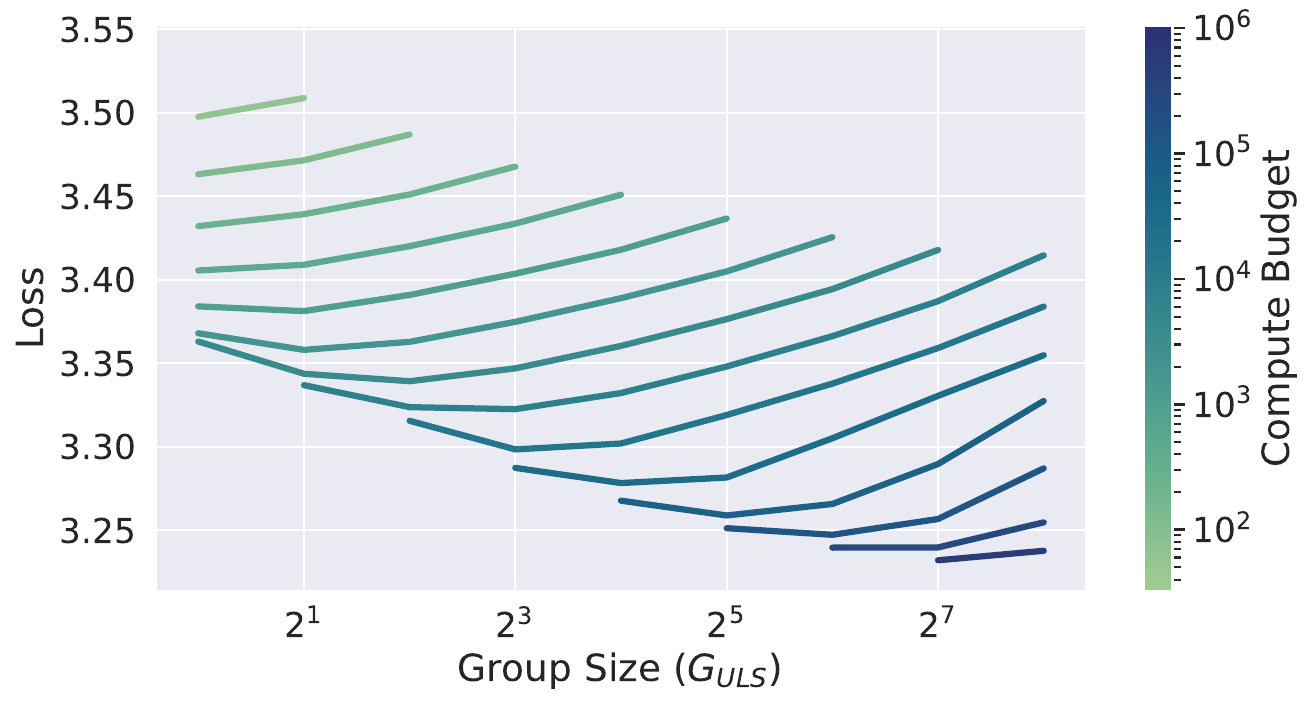}
        \caption{Loss for varying compute budgets and $G_\ulst$. Each curve represents a fixed compute budget.}
        \label{fig:uls-stackoverflow-group-size}
    \end{subfigure}
    \label{fig:uls-configure} 
    \caption{Loss of \uls on Stack Overflow for varying cohort and group sizes, with $\epsilon = 4$. Both plots are based on views of the same data, with the lines of (b) corresponding to the diagonals of (a).}
\end{figure}

The results for Stack Overflow with $\varepsilon = 4$ are in Figures \ref{fig:heatmap} and \ref{fig:uls-stackoverflow-group-size}. For a given compute budget $B = G_\ulst M$, it is clearly better to increase $M$ up to a limit (in this case 512) before increasing the $G_\ulst$ beyond $1$. For all compute budgets we used, the compute-optimal setting of $G_{\ulst}$ is always at least an order of magnitude smaller than $M$. By contrast, in federated learning, communication bottlenecks necessitate the use of larger $G_\ulst$.  For example, prior (federated) experiments on Stack Overflow \cite{denisov2022improved,kairouz2021practical,choquette2022multi,choquette2024amplified} set $G_{\ulst}=256$ and $M = 1000$, which is far from optimal in our setting; In data-center settings, we get lower loss for approximately the same compute budget by instead setting $G_{\ulst} = 64$, $M = 4096$.

Recall that by \eqref{eq:uls_gradient}, the noise variance is proportional to $L_\ulst \sigma_\elst$, where $L_\ulst$ is a the per-user gradient Lipschitz constant, and $\sigma_\ulst$ is the noise multiplier. Increasing $G_\ulst$ reduces $L_\ulst$ by a dataset-dependent factor, while increasing $M$ decreases $\sigma_\ulst$. While $\sigma_\ulst$ is easily computable via \eqref{eq:DP-SGD-ULS-accounting}, $L_\elst$ can be estimated as follows. Sample a set $U$ of users. For each $u\in U$, sample a subset $D_u$ of their dataset of size (at most) $G_\ulst$, and compute $\rho_u = \||D_u|^{-1}\sum_{z \in D_u} \nabla f(\theta, z) \|$ where $\theta$ is the pre-trained model. We estimate $L_\ulst \approx \median(\{\rho_u\vert u \in U\})$;  while \eqref{eq:alpha_and_beta} suggests using $\max(\cdot)$, $\median(\cdot)$ works well in practice.

We now use the following heuristic for choosing $G_\ulst$ and $M$ for a desired compute budget $B$. Start with initial $G_\ulst, M$ such that $G_\ulst M < B$. For simplicity, let these all be powers of 2. We compare $L_\ulst(2G_\ulst)/L_\ulst(G_\ulst)$ and $\sigma_\ulst(2M)/\sigma_\ulst(M)$. That is, we see whether doubling $G_\ulst$ or $M$ will reduce stochastic variance more, and then double whichever reduces variance more. We repeat until $G_\ulst  M = B$.  As we show in \cref{appendix:configure_uls}, this strategy is close to optimal in our empirical study. In the rest of our experiments, we set $G_\ulst$ and $M$ according to this strategy.

\begin{figure}[htb]
    \centering
    \includegraphics[width=0.9\linewidth]{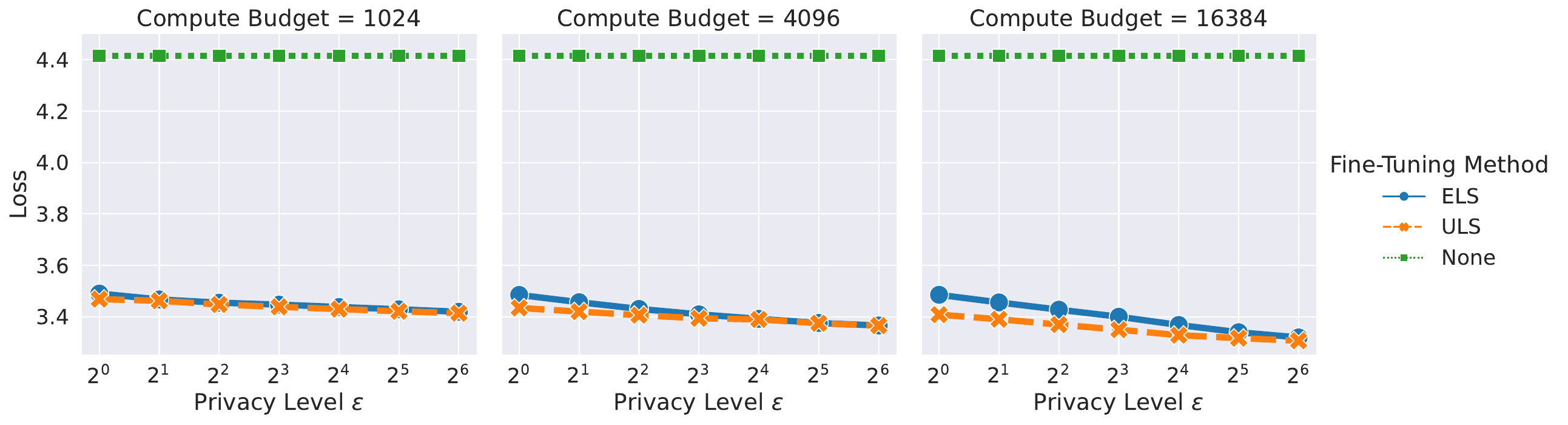}
    \caption{Privacy-loss trade-offs on Stack Overflow, for varying compute budgets.}
    \label{fig:loss_privacy_stackoverflow}
\end{figure}

\begin{figure}[htb]
    \centering
    \includegraphics[width=0.9\linewidth]{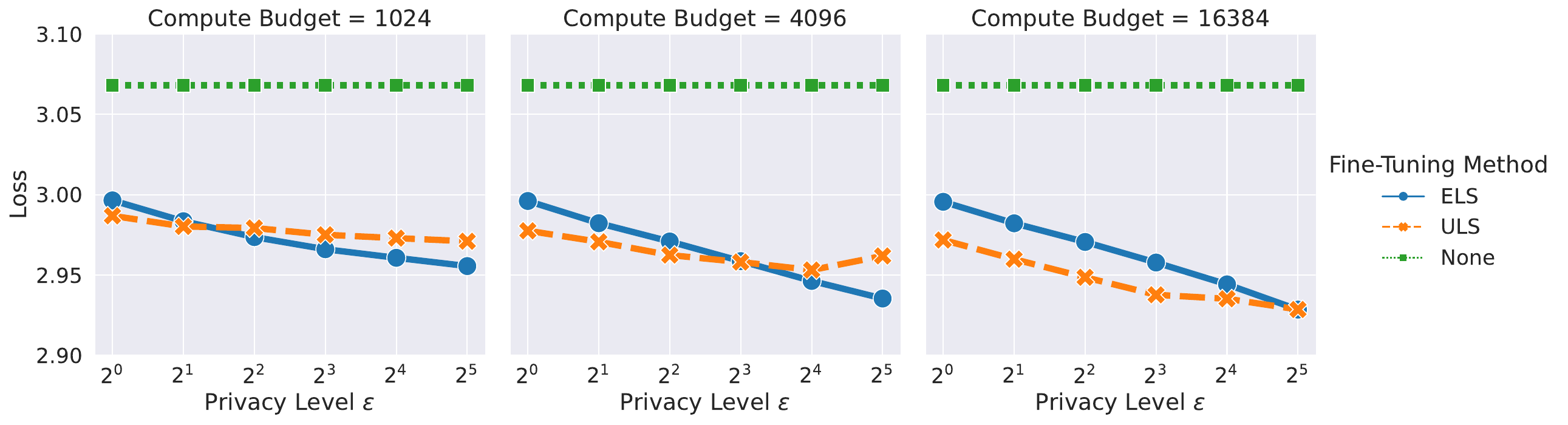}
    \caption{Privacy-loss trade-offs on CC-News, for varying compute budgets.}
    \label{fig:loss_privacy_ccnews}
\end{figure}

\mypar{Privacy-utility-compute trade-offs} We now apply \els and \uls for a variety of compute budgets and $\varepsilon$ values, using the heuristics above to select $G_\elst, G_\ulst$. We fine-tune on Stack Overflow and CC-News, and compute the test loss on the final iterate. In Figures \ref{fig:loss_privacy_stackoverflow} and \ref{fig:loss_privacy_ccnews}, we plot privacy-loss trade-offs for three distinct compute budgets. We also compare with using no fine-tuning whatsoever, that is, simply evaluating on the pre-trained checkpoint.

For Stack Overflow, \uls performs at least as well as \els in all settings. The magnitude of the improvement is largest for small $\varepsilon$ and large compute budgets, aligning with what we observed for simpler settings in Sections \ref{sec:understanding} and \ref{sec:synthetic_task}. For CC-News, \uls outperforms \els in many settings, especially when the compute budget is large or $\varepsilon$ is small. For both datasets, \uls improves more with increased compute budgets: if we fix a privacy level $\epsilon$, the loss for \uls drops more significantly as the compute budget increases than for \els. We demonstrate this in greater detail in \cref{appendix:compute_loss}. Notably, fine-tuning with user-level DP provides significant wins over simply using the pre-trained checkpoint in all settings.

\begin{figure}[htb]
    \centering
    \includegraphics[width=0.8\linewidth]{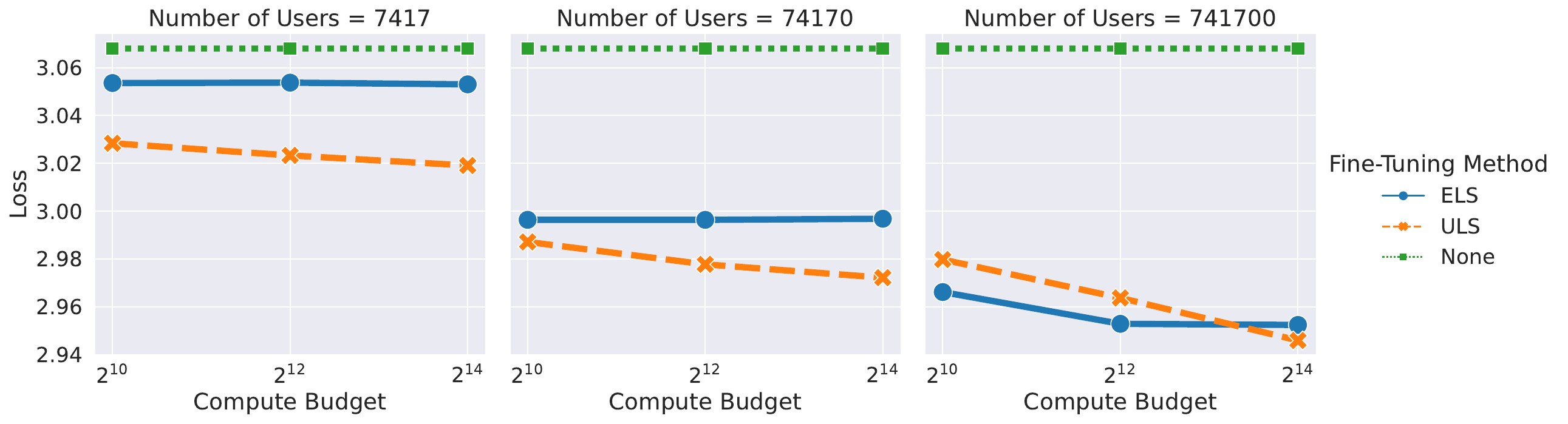}
    \caption{Compute versus loss on CC-News, as the number of users in the privacy accounting varies.}
    \label{fig:loss_budget_ccnews_num_users_ablation}
\end{figure}

\mypar{Impact of dataset size} We investigate, for a fixed compute budget, how \els and \uls compare as we vary the number of users. To test this, we use CC-News. We fix all other factors, but vary the number of users \emph{only} in the privacy accounting. Recall from Section \ref{sec:exp-setup} that our results above on CC-News assume there are $10\times$ more users than are actually present, for the accounting. We perform the same experiments, but instead assume there are $1\times, 10\times,$ and $100\times$ more users in the dataset in the accounting. The results for $\varepsilon = 4$ are in Figure \ref{fig:loss_budget_ccnews_num_users_ablation}. Throughout, the loss of \uls decreases more as compute budget increases, but the loss of \els decreases more as the number of users increases.

\mypar{Model personalization} In some settings, we may wish to personalize the model after fine-tuning on user data. We measure the ability of the fine-tuned models to personalize to downstream user data. We find that both algorithms seem to benefit comparably from personalization, which does not change the general trade-offs discussed above. For details and results, see Appendix \ref{appendix:personalization}.

\section{Related Work}\label{sec:related_work}

\subsection{Comparison to \citet{chua2024mindprivacyunituserlevel}}

Concurrently with our work, \citet{chua2024mindprivacyunituserlevel} studied the same algorithms for fine-tuning LLMs with user-level differential privacy. As in our work, they empirically compare the effectiveness of \els and \uls in datacenter settings. In comparison to our work, \cite{chua2024mindprivacyunituserlevel} focuses on LoRA fine-tuning, compares \els and \uls to the algorithm of \cite{liu2024user}, studies the sensitivity of the algorithms to clipping norms, and empirically analyzes various data selection strategies within users. On the other hand, our work expands upon the work of \cite{chua2024mindprivacyunituserlevel} in a few key ways. First, our new accounting techniques enable fairer comparisons between \els and \uls. Second, the largest dataset \cite{chua2024mindprivacyunituserlevel} uses has 240,173 examples partitioned across approximately 18,000 users, while our experiments encompass a dataset (Stack Overflow) with 135,818,730 examples partitioned across 342,477 users. Third, our comparisons between \els and \uls normalize on total compute budget, giving us a principled way to compare \els and \uls as the compute budget changes, which can significantly impact their relative comparison. Last, we provide explicit methodology for selecting the group size in both \els and \uls, and give extensive empirical evidence for the utility of these methods.

\subsection{Theoretical Advances in User-Level DP}

Here, we survey the existing theoretical work on user-level DP in various settings. 

\mypar{Reductions to example-level DP}
\citet{ghazi2023user1,ghazi2023user} give user-level DP bounds by running example-level DP algorithms on subsets of data where the algorithm is stable to deletions (\cite{ghazi2023user1} only handles output perturbation while \cite{ghazi2023user} can handle any arbitrary algorithm).
This involves a brute-force-search for a stable subset of the data (together with a propose-test-release loop), which can be super-polynomial in the number of examples, and is thus computationally intractable. We note that the user-level DP bounds of \cite{ghazi2023user} rely on generic group privacy reductions~\cite{vadhan2017complexity}. In contrast, we give tight accounting for deriving user-level DP guarantees from example-level DP guarantees in the specific setting of \elsfull, enabling us to scale this method to practical LLM settings.

We also note that prior works~\cite{amin2019bounding,epasto2020smoothly} have also reduced learning with user-level DP to the example-level DP setting where each example is associated with a weight, and the weights are computed to maximize the final utility. 
On the other hand, we fix a number $G_\elst$ of examples per user globally and randomly select $G_\elst$ samples; this can be viewed as assigning a binary per-example weights.
It is unclear how to adapt the analytical approaches of~\cite{amin2019bounding,epasto2020smoothly} to LLMs as we do not have access to utility bounds.

\mypar{Approaches based on clipping user-level gradients}
Several prior approaches rely on bounding user contributions by clipping user-level gradients and combining them with (different choices of) robust mean estimation algorithms~\cite{levy2021learning,bassily2023user,liu2024user,de2022unlocking}. This line of work aims for better rates of convergence under the weaker assumptions. On the other hand, we empirically evaluate practical algorithms derived from these approaches by using user-level gradient clipping but replacing the inefficient robust aggregation approaches with a simple unweighted average (which is efficiently implementable on hardware accelerators).

\mypar{Simpler theoretical settings}
Related work also considers user-level DP in stylized problems such as learning discrete distributions \citep{liu2020learning} and histograms \citep{liu2023algorithms}. Mean estimation under user-level DP was also considered in \cite{girgis2022distributed,narayanan2022tight,cummings2022mean}. This subroutine was also the key building block of the theoretical works~\cite{levy2021learning,bassily2023user,liu2024user} in their learning bounds under user-level DP.
Continual mean estimation~\cite{george2024continual} and
continual counting~\cite{dong2023continual} have also been considered under user-level DP.

\mypar{Other related work}
\citet{fang2024privacy} give an approach to privacy amplification from sampling users in a graph structure.

\subsection{Federated Learning and User-Level DP}
As we noted in the main paper, empirical advances in user-level DP have been driven primarily by research in federated learning, starting with \citep{geyer2017dp,mcmahan2018learning,agarwal2018cpsgd,kato2023uldp}. Specifically, \citet{mcmahan2018learning} propose DP-FedSGD (resp. DP-FedAvg) which clips user-level gradients (resp. pseudo-gradients generated by multiple local gradient steps). Note that DP-FedSGD coincides with \ulsfull that we analyze. This user-level gradient clipping approach can be generalized beyond DP-SGD to algorithms that add noise that is correlated across iterations~\cite{kairouz2021practical}; these algorithms have been deployed in industrial systems to provide formal user-level DP guarantees~\cite{xu2023federated}.

\citet{kato2023uldp} give user-level DP algorithms in a cross-silo federated learning setting. Here, a user's data might be split across multiple data silos and each silo might contain multiple datapoints from a single user. Their approach combines \elsfull with FedAvg, where they use generic group privacy reductions to promote example-level DP guarantees to the user-level.

Several follow-up works have leveraged algorithms similar to \ulsfull in other settings such as contextual bandits~\cite{huang2023federated}, meta-learning~\cite{li2020dp,zhou2022task}, and embedding learning~\cite{xu2023learning} as well as applications such as medical image analysis~\cite{adnan2022federated}, speech recognition~\cite{pelikan2023federated} and releasing mobility reports (i.e. aggregate location statistics)~\cite{kapp2023towards}.

\section{Discussion}\label{sec:discussion}

Our work highlights two general findings. First, by introducing tight group-level accounting, we can make \els a practical method for user-level DP that is scalable to LLM settings and serves as a useful baseline for subsequent research. Second, despite this improvement in accounting, across a wide variety of settings (synthetic and realistic) \uls often outperforms \els. While we are able to scale \uls to models with hundreds of millions of parameters, \uls uses user-level sampling that is distinct from most LLM training algorithms. Future work is needed to determine which algorithms best balance scalability and performance when training with formal user-level DP guarantees.

\mypar{Limitations} Despite theoretical advancements in understanding \els and \uls, we do not have a general theorem comparing their utility across different settings. We also do not provide theoretical comparisons in the absence of Lipschitz loss functions. Empirically, both \els and \uls depend crucially on group sizes $G_\elst$ and $G_\ulst$. While we provide practical heuristics that work well across experiments (\cref{sec:exp-results}), it is not obvious that these will carry over to all settings of interest. Additionally, due to compute constraints, our work focused on exploring a breadth of hyperparameter configurations rather than on maximally scaling up model size. Determining the model scaling behavior of \els and \uls is an important topic for future work.

\bibliographystyle{unsrtnat}
\bibliography{references}

\appendix

\subsection{User-level Privacy Attacks}

The work on user-level DP is motivated in part due to the various privacy attacks conducted at the user level. For instance, an adversary might be able to infer whether a user's data was used to train a model~\cite{song2019auditing}, even if the adversary does not have access to the exact training samples of the user~\cite{kandpal2023user}.
Further, \cite{kandpal2023user} demonstrate that example-level DP is not effective in mitigating such user inference attacks, especially at low false positive rates.

Such attacks have been designed not only for LLMs~\cite{kandpal2023user} but also for embedding learning for vision~\cite{li2022userlevel}, speech recognition for IoT devices~\cite{miao2021audio}, facial recognition systems~\citep{face_auditor}. In federated learning, \citet{Wang_FL_User_2019} and \citet{Song_FL_User_Privacy_2020} study the risk to user-level privacy from a malicious server.
Such privacy attacks can also be used to audit user-level DP~\cite{jagielski2020auditing,pillutla2023unleashing,steinke2023privacy} and estimate user-level privacy~\cite{andrew2024one}.
User-level DP provides formal upper bounds on the success rates of all such user-level privacy attacks and the algorithms we study are broadly applicable in all of these settings.

\section{Tight Accounting for \elsfull}\label{sec:accounting}

\mypar{Notation} 
Given a distribution $P$ over a space $\calA$ and $n \in \mathbb{Z}_{> 0}$, let $P^{\otimes n}$ denote the product distribution on $\calA^n$.

In this section, we show the following tight accounting statement for \cref{alg:dp-sgd-els}:

\begin{thm}\label{thm:dp-sgd-accounting}
Let $\calM_f$ be $T$ iterations of \elsfull using noise multiplier $\sigma$, loss function $f$, group size $G_\elst = K$, and Poisson sampling with probability $p$. That is, $\calM_f(D)$ is the distribution of models $(\theta^1, \theta^2, \ldots \theta^T)$ produced by applying \elsfull to a dataset $D$. Let $D$ and $D'$ be two datasets such that $D' = D \sqcup A$ where $|A| \leq K$. Then for all $\varepsilon$:

\[H^{\sym}_{e^\varepsilon}(\calM_f(D), \calM_f(D')) \leq H^{\sym}_{e^\varepsilon}(\calN(0, \sigma^2)^{\otimes T}, \calN(\bin(K, p), \sigma^2)^{\otimes T}).\]

Furthermore, this is tight, i.e. there exists a loss function $f$ and datasets $D, D'$ such that:

\[H^{\sym}_{e^\varepsilon}(\calM_f(D), \calM_f(D')) = H^{\sym}_{e^\varepsilon}(\calN(0, \sigma^2)^{\otimes T}, \calN(\bin(K, p), \sigma^2)^{\otimes T}).\]
\end{thm}

Since we cap the number of examples any user can contribute to the dataset in Algorithm~\ref{alg:dp-sgd-els}, this implies Theorem~\ref{thm:dp-sgd-accounting-informal}. To prove this, we use the following lemma from \cite{choquettechoo2024privacy}, derived using an analysis based on Mixture-of-Gaussians mechanisms:

\begin{lem}[Lemma 4.5 of \cite{choquettechoo2024privacy}]\label{lem:vector-to-scalar-reduction}
Let $\boldx$ be a random variable on $\R^d$, and $x$ be a random variable on $\R$ such that $\ltwo{\boldx}$ is stochastically dominated by $x$ (that is, there is a coupling of $\boldx$ and $x$ such that under this coupling, $\ltwo{\boldx} \leq x$ with probability 1). Then for all $\varepsilon > 0$:

\[H_{e^\varepsilon}(\calN(0, \sigma^2I_d), \calN(\boldx, \sigma^2 I_d)) \leq H_{e^\varepsilon}(\calN(0, \sigma^2), \calN(x, \sigma^2)),\]
\[H_{e^\varepsilon}(\calN(\boldx, \sigma^2 I_d), \calN(0, \sigma^2I_d)) \leq H_{e^\varepsilon}(\calN(x, \sigma^2), \calN(0, \sigma^2)).\]
\end{lem}

We will also use the following ``quasi-convexity'' property of DP:

\begin{lem}\label{lem:quasi-convexity}
Let $w_1, w_2, \ldots, w_n \geq 0$ be probabilities summing to 1. Given distributions $\{P_i\}, \{Q_i\}$, let $P = \sum_i w_i P_i$ and $Q = \sum_i w_i Q_i$. Then for any $\alpha \geq 0$:

\[H_\alpha(P, Q) \leq \max_i H_\alpha(P_i, Q_i).\]
\end{lem}
\begin{proof}
We have:

\begin{align*}
H_\alpha(P, Q) &= \int \max\{P(x) - \alpha Q(x), 0\} dx = \int \max\{\sum_i w_i (P_i(x) - \alpha Q_i(x)), 0\} dx\\
&\stackrel{(\ast_1)}{\leq} \int \sum_i w_i \max\{P_i(x) - \alpha Q_i(x), 0\} dx = \sum_i w_i \int \max\{P_i(x) - \alpha Q_i(x), 0\} dx\\
&= \sum_i w_i H_\alpha(P_i, Q_i) \stackrel{(\ast_2)}{\leq} \max_i H_\alpha(P_i, Q_i).\\
\end{align*}

$(\ast_1)$ is the observation that $\max\{a+c, b+d\} \leq \max\{a, b\} + \max\{c, d\}$ i.e. ``the max of sums is less than the sum of maxes'' and $(\ast_2)$ holds because the $w_i$ are non-negative and sum to 1.
\end{proof}

Finally, we will use the following observation about bijections and hockey-stick divergences:

\begin{obs}
Let $f$ be any bijection, and for distribution $X$ let $f(X)$ denote the distribution given by $f(x), x \sim X$. Then for any $P, Q, \varepsilon$:
\[H^{\sym}_{e^\varepsilon}(P, Q) = H^{\sym}_{e^\varepsilon}(f(P), f(Q)).\]
\end{obs}
\begin{proof}
This follows by applying the post-processing property of DP, which says for any function $g$

\[H^{\sym}_{e^\varepsilon}(P, Q) \geq H^{\sym}_{e^\varepsilon}(g(P), g(Q)).\]

The observation follows by applying the post-processing property to $f$ and $f^{-1}$.
\end{proof}

\begin{proof}[Proof of Theorem~\ref{thm:dp-sgd-accounting}]

Since $(\theta^1, \theta^2, \ldots, \theta^T) \leftrightarrow (\theta^1 - \theta^0, \theta^2 - \theta^1, \ldots, \theta^T - \theta^{T-1})$ is a bijection (we treat the initialization $\theta^0$ as public), we can assume through the proof that \elsfull instead outputs the tuple $(\theta^1 - \theta^0, \theta^2 - \theta^1, \ldots, \theta^T - \theta^{T-1})$. For simplicity of presentation, we will assume $f$ is $C$-Lipschitz and thus that $\clip$ is a no-op.

The tightness of this results follows from considering a one-dimensional 1-Lipschitz loss function $f$ such that that for all $z \in D$, $f(\theta, z) = 0$ and for all $z \in A$, $f(\theta, z) = -\theta$. Letting $\eta = 1$, the distribution of each $\theta^t - \theta^{t-1}$ is exactly $x \sim \calN(0, \sigma^2)$ for $D$ and $x \sim \calN(\bin(K, p), \sigma^2)$ for $D'$. 

For the upper bound, we will show 
\[H_{e^\varepsilon}(\calM_f(D), \calM_f(D')) \leq H_{e^\varepsilon}(\calN(0, \sigma^2)^{\otimes T}, \calN(\bin(K, p), \sigma^2)^{\otimes T}).\] 

The analogous bound on $H_{e^\varepsilon}(\calM_f(D'), \calM_f(D))$ (and thus the desired bound on $H^{\sym}_{e^\varepsilon}(\calM_f(D), \calM_f(D'))$) follows by Lemma 28 of \cite{zhu22optimal}.

By adaptive composition of privacy loss distributions (see e.g. Theorem 2.4 of \cite{doroshenko22connect}), it suffices to show given any fixed $\theta^t$, if $P, Q$ are the distribution of $\theta^{t+1} - \theta^t$ conditioned on $\theta^t$ using $D$ and $D'$ respectively, then for all $\varepsilon$ we have:

\[H_{e^\varepsilon}(P, Q) \leq  H_{e^\varepsilon}(\calN(0, \sigma^2), \calN(\bin(K, p), \sigma^2)).\]

Recall that $S^{t+1}$ is the set of examples sampled in iteration $t$,  and let $P_S, Q_S$ denote the distributions of $P, Q$ respectively conditioned on the event $S^{t+1} \cap D = S$. The distribution of $S^{t+1} \cap D$ is the same for $D$ and $D'$, so by Lemma~\ref{lem:quasi-convexity} for all $\varepsilon$:

\[H_{e^\varepsilon}(P, Q) \leq \max_{S} H_{e^\varepsilon}(P_S, Q_S),\]

hence it suffices to show for any fixed $S$ and all $\varepsilon$:

\[\max_{S} H_{e^\varepsilon}(P_S, Q_S) \leq  H_{e^\varepsilon}(\calN(0, \sigma^2), \calN(\bin(K, p), \sigma^2)).\]

Now let $P_S' = -\frac{p + \eta \sum_{z \in S} \nabla f(\theta^t, z) }{\eta C}$ where $p \sim P_S$. We define $Q_S'$ analogously. The correspondence
\[
p \leftrightarrow -\frac{p + \eta \sum_{z \in S} \nabla f(\theta^t, z) }{\eta C}
\]
is a bijection on $\mathbb{R}^d$, so $H_{e^\varepsilon}(P_S, Q_S) = H_{e^\varepsilon}(P_S', Q_S')$. We can exactly write the distributions of $P_S', Q_S'$:

\[P_S' = \calN(0, \sigma^2 I_d), Q_S' = \calN\left(\sum_{z \in S^{t+1} \cap A} \frac{\nabla f(\theta^t, z)}{C}, \sigma^2 I_d\right).\]

By triangle inequality and $C$-Lipschitzness of $f$, $\ltwo{\sum_{z \in S^{t+1} \cap A} \frac{\nabla f(\theta^t, z)}{C}} \leq |S^{t+1} \cap A|$. Furthermore, $|S^{t+1} \cap A|$ is distributed according to $\bin(|A|, p)$, and so $\ltwo{\sum_{z \in S^{t+1} \cap A} \frac{\nabla f(\theta^t, z)}{C}}$ is stochastically dominated by $\bin(K, p)$. By Lemma~\ref{lem:vector-to-scalar-reduction} we now have for all $\varepsilon$:

\[H_{e^\varepsilon}(P_S', Q_S') \leq H_{e^\varepsilon}(\calN(0, \sigma^2), \calN(\bin(K, p), \sigma^2)),\]

which completes the proof.
\end{proof}

\subsection{Implementation in \texttt{dp\_accounting}}\label{sec:accounting-implementation}

\subsubsection{Computing $\varepsilon$ and $\delta$}

The following code snippet using the \texttt{dp\_accounting} library~\cite{dp_accounting} and \texttt{scipy} methods can be used to compute $\varepsilon$ as a function of $\delta$ (or vice-versa) for \elsfull (Algorithm~\ref{alg:dp-sgd-els}) according to Theorem~\ref{thm:dp-sgd-accounting-informal}, for a given number of steps $T$, example sampling probability $p$, noise multiplier $\sigma_\elst =\sigma$, and group size $G_\elst = K$:

\begin{verbatim}
def get_group_level_event(T, p, sigma, K):
    sensitivities = range(K+1)
    probs = [scipy.stats.binom.pmf(x, K, p) for x in sensitivities]
    single_round_event = dp_accounting.dp_event.MixtureOfGaussiansDpEvent(
      sigma, sensitivities, probs
    )
    dp_sgd_event = dp_accounting.dp_event.SelfComposedDpEvent(
      single_round_event, T
    )
    return dp_sgd_event

event = get_group_level_event(T, p, sigma, K)
accountant = dp_accounting.pld.PLDAccountant()
accountant.compose(dp_sgd_event)

# Compute epsilon given delta
print(accountant.get_epsilon(delta))

# Compute delta given epsilon
print(accountant.get_delta(epsilon))
\end{verbatim}

\subsubsection{Computing $\sigma_\elst$}

To figure out the minimum $\sigma_\elst$ needed to achieve a target $(\varepsilon, \delta)$-DP guarantee for \elsfull (\cref{alg:dp-sgd-els}), we can use \texttt{dp\_accounting}'s \texttt{calibrate\_dp\_mechanism}:

\begin{verbatim}
def get_group_level_sigma(T, p, epsilon, delta, K)
    sigma_to_event = lambda sigma: get_group_level_event(T, p, sigma, K)
    return dp_accounting.calibrate_dp_mechanism(
        dp_accounting.pld.PLDAccountant,
        sigma_to_event,
        epsilon,
        delta
    )
\end{verbatim}

\section{Comparing Variances of \els and \uls}\label{appendix:compare_variances}

Recall that in the setting of \cref{sec:understanding}, the stochastic gradients produced by \els and \uls in an iteration $t$ are respectively given by
\begin{align*}
g_{\elst}^t &= \dfrac{1}{B} \sum_{z \in S^t} \nabla f(\theta^t, z)  + \zeta_{\elst}^t,~~~\zeta_{\elst}^t \sim \mathcal{N}\left(0, \left(\dfrac{\sigma_{\elst}L_{\elst}}{B}\right)^2I_d\right)\,, \\
g_{\ulst}^t &= \dfrac{1}{B} \sum_{u\in U^t} \sum_{z \in D_u} \nabla f(\theta^t, z) + \zeta_{\ulst}^t,~~~\zeta_{\ulst}^t \sim \mathcal{N}\left(0, \left(\dfrac{\sigma_{\ulst}L_{\ulst}}{M}\right)^2I_d\right)\,,
\end{align*}
where $B$ denotes the per-iterate (expected) compute budget of both methods, and $M$ is the (expected) cohort size of \uls. Further recall that in this setting, there are $N$ users, each of which have $K$ examples.

For any specific instatiation of this setting, we can use the DP accounting tools from \cref{sec:accounting} to explicitly compute $\variance(\zeta_\elst^t)$ and $\variance(\zeta_\ulst^t)$.

We do so in the following setting. We fix $N = 1024$ users, each with $K = 32$ examples. We set $G_\elst = 32$ (though the choice here has almost no impact on the noise variance). We vary the cohort size $M$, and set $G_\ulst = B/M$ to normalize compute between \els and \uls. We fix $T = 1000, \delta = 10^{-6}$ and vary $\epsilon$ in the desired $(\epsilon, \delta)$-DP guarantee, and compute the corresponding noise multipliers $\sigma_\elst, \sigma_\ulst$ via DP accountants.

To compute variance, the only remaining relevant quantities are $L_\elst$ and $L_\ulst$. We fix $L_\elst = 10$, and vary $L_\ulst$. Intuitively, the ratio of these two tells us how diverse the gradients across a user are. We consider two settings. In the first, the $G_\ulst$ gradients across a user in \uls are minimally diverse, so that $L_\elst = L_\ulst$ for all group sizes $G_\ulst$. In the second, the gradients are maximally diverse, so that
\[
L_\ulst = \dfrac{L_\elst}{\sqrt{G_\ulst}}.
\]
This setting occurs, for example, if all $G_\ulst$ gradients computed at each user for \uls are orthogonal with length $L_\elst$. For varying $B, M$ and $\epsilon$, we then compare three variances: the variance of $\zeta_\els^t$, and the variance of $\zeta_\ulst^t$ for each setting of $L_\ulst$.

\begin{figure}[htb]
    \centering
    \includegraphics[width=\linewidth]{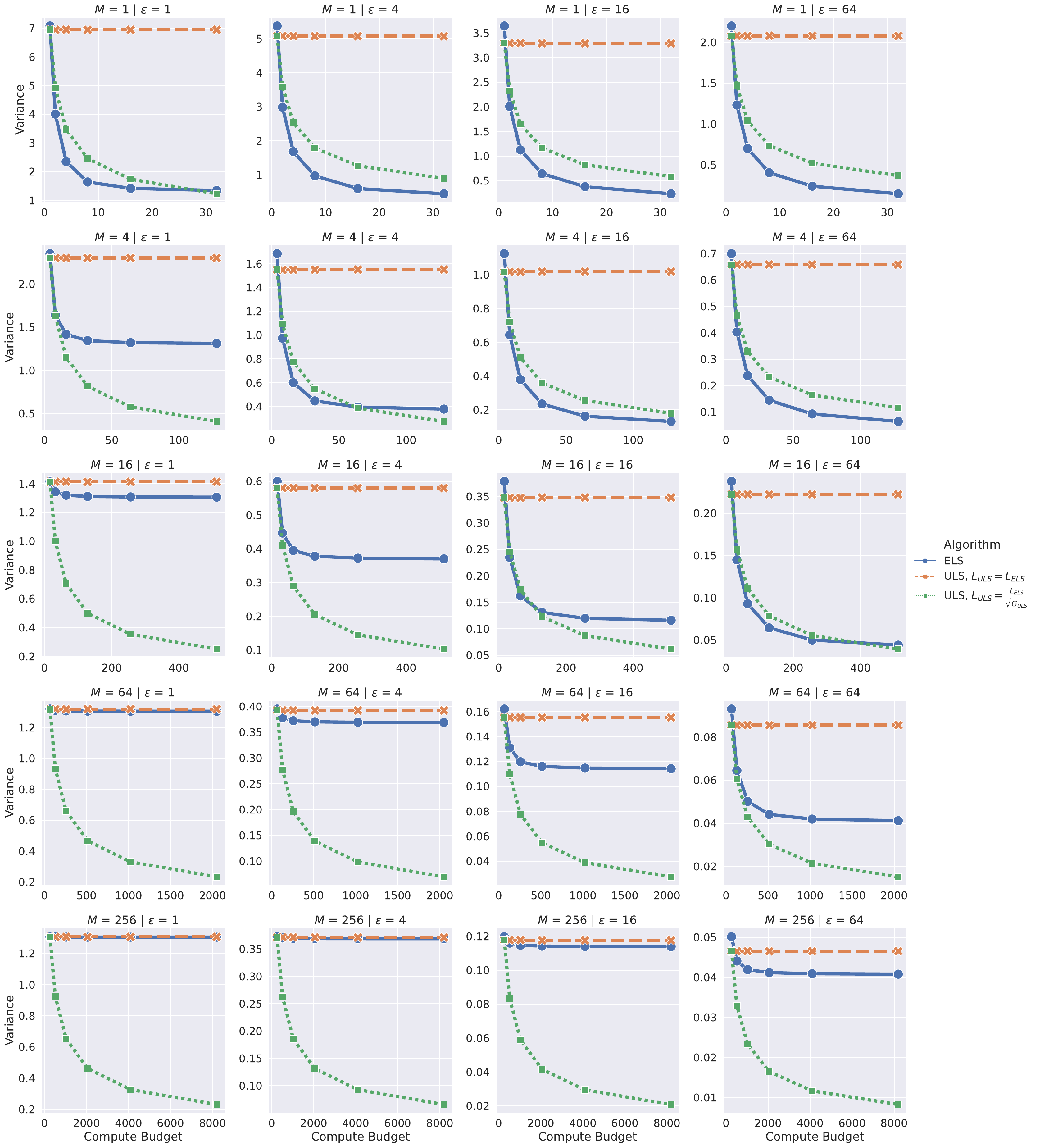}
    \caption{\small 
    Noise variance of \els and \uls, for varying compute budget $B$, cohort size $M$ and privacy level $\epsilon$. For \uls, we fix the cohort size $M$ and vary $G_\ulst$. We compare two settings, one in which $L_\ulst = L_\elst$, and one in which $L_\ulst = L_\elst / \sqrt{G_\ulst}$. Throughout, we fix $N = 1024$ users, each with $K = 32$ examples, $G_\elst = 32$, $T = 1000$, $\delta = 10^{-6}$, and $L_\els = 10$.}
    \label{fig:compare_variance_full}
\end{figure}

The results are given in \cref{fig:compare_variance_full}. While the results vary across settings, we see a few robust findings. First, when $L_\ulst = L_\elst$, the variance of \els is lower in nearly all settings. When $L_\ulst = L_\elst / \sqrt{G_\ulst}$, the variance of \uls is often (but not always) lower than that of \els. We see that \uls especially tends to incur lower variance when either (1) $\epsilon$ is small or (2) when the compute budget is sufficiently large.

\clearpage

\section{Proof of Lemma~\ref{lem:conj-ub}}\label{sec:conj-ub-proof}

\begin{proof}

Let $B_{K, p}(c) = \Pr[\bin(K, p) = c]$. By linearity of expectation, we have
\allowdisplaybreaks
\begin{align*}
    &\exp((\alpha - 1) R_\alpha(P_K(K \sigma), Q(K \sigma)))\\ &= \mathbb{E}_{x \sim \calN(0, K^2 \sigma^2)}\left[\left(\sum_{c \in \{0, 1, \ldots, K\}} B_{K,p}(c) \cdot \exp\left( \frac{2c x -c^2}{2K^2\sigma^2}\right)\right)^\alpha\right]\\
    &= \mathbb{E}_{x \sim \calN(0, K^2 \sigma^2)}\left[\sum_{c_1, c_2, \ldots, c_\alpha \in \{0, 1, \ldots, K\}} \left(\prod_i B_{K,p}(c_i)\right) \cdot \exp\left( \frac{2 (\sum_i c_i) x -\sum_i c_i^2}{2K^2\sigma^2}\right)\right]\\
    &= \sum_{c_1, c_2, \ldots, c_\alpha \in \{0, 1, \ldots, K\}} \left(\prod_i B_{K,p}(c_i)\right) \cdot \mathbb{E}_{x \sim \calN(0, K^2 \sigma^2)}\left[\exp\left( \frac{2 (\sum_i c_i) x -\sum_i c_i^2}{2K^2\sigma^2}\right)\right]\\
    &= \sum_{c_1, c_2, \ldots, c_\alpha \in \{0, 1, \ldots, K\}} \left(\prod_i B_{K,p}(c_i)\right) \cdot \exp\left( \frac{(\sum_i c_i)^2 -\sum_i c_i^2}{2K^2\sigma^2}\right)\end{align*}

This last step follows from the fact that for $a, \nu \in\R$ and $y \sim \calN(0, \nu^2)$, $\E_{y}[e^{ay/\nu^2}] = e^{a^2 / 2 \nu^2}$. For $c_i \sim \bin(K, p)$, we define random variables $\{c_{i, j}\vert j \in \{0, 1, \ldots, K\}, c_{i, j}\sim\bern(p)\}$, and can write $c_i = \sum_{j\in \{0, 1, \ldots, K\}}c_{i,j}$. Let $C = (c_{1, 1}, \dots, c_{\alpha, K}) \sim \bern(p)^{\alpha K}$. Then we have:

\begin{align*}
& \sum_{c_1, c_2, \ldots, c_\alpha \in \{0, 1, \ldots, K\}} \left(\prod_i B_{K,p}(c_i)\right) \cdot \exp\left( \frac{(\sum_i c_i)^2 -\sum_i c_i^2}{2K^2\sigma^2}\right) \\
    &= \mathbb{E}_{C}\left[ \exp\left( \frac{(\sum_{i,j} c_{i,j})^2  -\sum_i (\sum_j c_{i,j})^2}{2K^2\sigma^2}\right)\right]\\
    &= \mathbb{E}_{C}\left[ \exp\left( \frac{\sum_{i \neq i', j, j'} c_{i,j} c_{i', j'}}{2K^2\sigma^2}\right)\right]\\
    &= \mathbb{E}_{C}\left[\exp\left( \frac{\mathbb{E}_{j_1, j_2, \ldots j_\alpha \stackrel{u.a.r.}{\sim}\{0, 1, \ldots, K\}} [\sum_{i \neq i'} c_{i,j_i}c_{i',j_{i'}}]}{2\sigma^2}\right)\right]\\
    \qquad\text{(by Jensen's inequality)}&\leq \mathbb{E}_{C}\left[\mathbb{E}_{j_1, j_2, \ldots j_\alpha \stackrel{u.a.r.}{\sim}\{0, 1, \ldots, K\}} \left[\exp\left( \frac{\sum_{i \neq i'} c_{i,j_i}c_{i',j_{i'}}}{2\sigma^2}\right)\right]\right]\\
    \qquad\text{(by the law of total expectation)} &= \mathbb{E}_{c_1, \ldots, c_\alpha \sim \bern(p)}\left[ \exp\left( \frac{\sum_{i \neq i'} c_i c_{i'}}{2\sigma^2}\right)\right]\\
    &=\exp((\alpha-1)R_\alpha(P_1(\sigma), Q(\sigma))).
\end{align*}
\end{proof}

\clearpage

\section{Datasets}\label{appendix:datasets}

\begin{figure}[htb]
    \centering
    \begin{subfigure}{0.4\linewidth}
        \includegraphics[width=\linewidth]{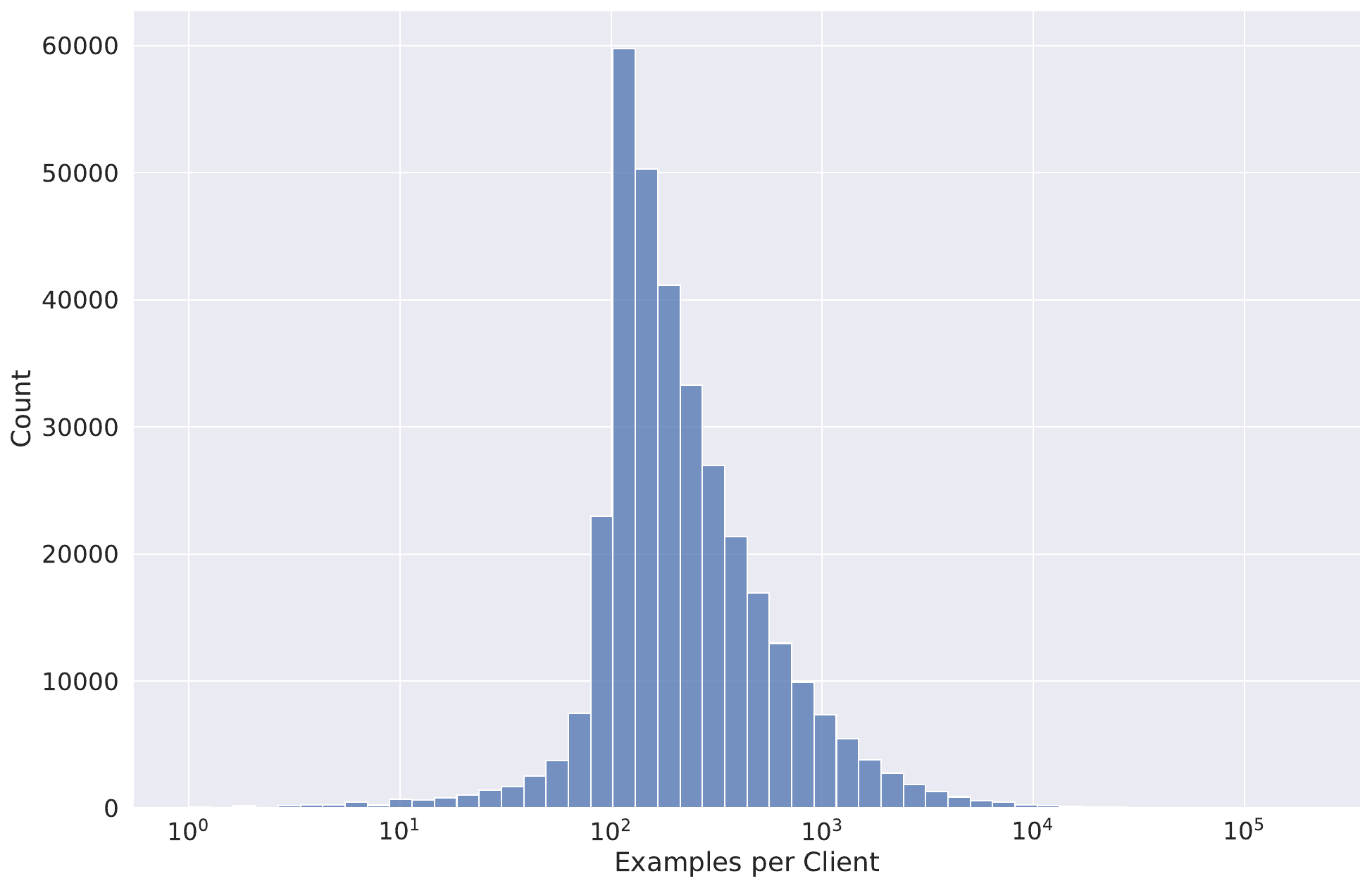}
    \end{subfigure}
    \begin{subfigure}{0.4\linewidth}
         \includegraphics[width=\linewidth]{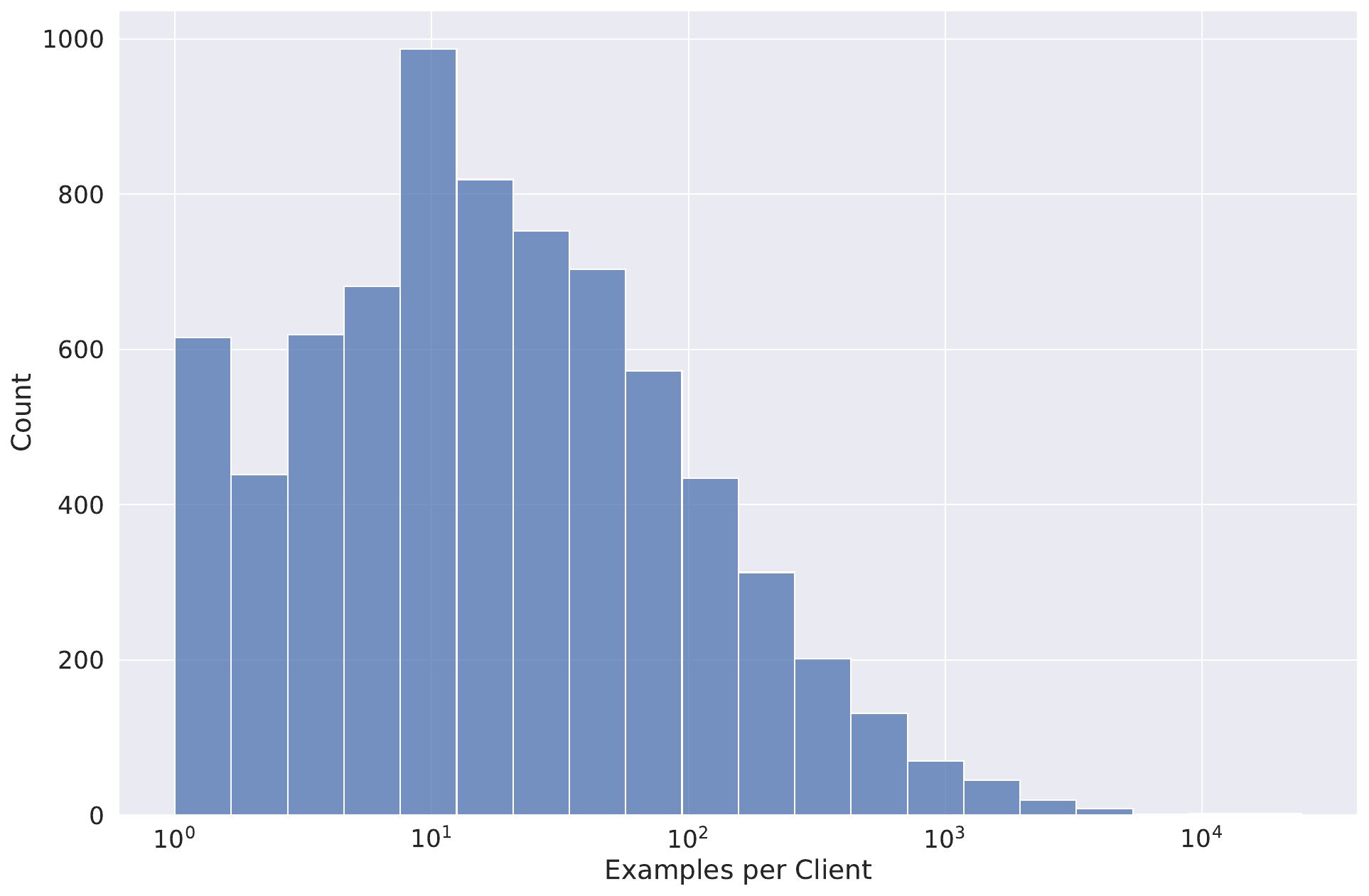}
    \end{subfigure}
    \caption{Histograms of user dataset sizes for the training splits of Stack Overflow (left) and CC-News (right).}
    \label{fig:dataset_histograms}
\end{figure}

We use two user-partitioned datasets, Stack Overflow and CC-News, for fine-tuning and evaluation. The Stack Overflow dataset \citep{stackoverflow}, consists of questions and answers from \url{stackoverflow.com}, and is naturally partitioned by user on the platform. The dataset contains three splits: train (examples before 2018-01-01 UTC), test (examples after 2018-01-01 UTC) and validation (examples from held-out users across time). For our experiments, we use the train split for fine-tuning and the test split for evaluation. \cref{fig:dataset_histograms} depicts the distribution of user dataset size in the train split. Stack Overflow evaluation metrics reported throughout the main paper are measured on the full test split. In \cref{appendix:personalization}, we also include an ablation on model personalization using Stack Overflow test. Personalization is done by using half of each test user's examples for further fine-tuning and evaluating each personalized model on the reserved half of each test user's examples. 

CC-News consists of English-language articles on the web, a subset of the Colossal Clean Crawled Corpus (C4) dataset. For CC-News, we leverage Dataset Grouper to obtain user-level partitioning by base domain of each article's URL~\citep{charles2023towards}. We reserve a portion of each user's data in CC-News for evaluation and train on the remainder. In order to do so, we remove all users with only a single example. Eval user datasets are then formed by taking the first 10\% of the data of each user, but only up to a maximum of 32 examples. The remaining 90\% of user data is allocated for training. \cref{fig:dataset_histograms} shows the distribution of examples per user in the CC-News training dataset, after the held-out portion of examples are removed. Dataset statistics for Stack Overflow and CC-News are reported in \cref{dataset_stats}.

\begin{table}[ht]
\caption{User-level statistics for the train and test splits of Stack Overflow and CC-News.}
\label{dataset_stats}
\centering
\begin{tabular}{@{}llccccc@{}}
\toprule
\multirow{2}{*}{\textbf{Dataset}}        & \multirow{2}{*}{\textbf{Split}} & \multicolumn{2}{c}{\textbf{Dataset-level Statistics}} & \multicolumn{3}{c}{\textbf{User-level Statistics (\# examples / user)}} \\ \cmidrule(lr){3-4}\cmidrule(lr){5-7} 
&& \# Examples & \# Users & Min & Median & Max \\ \midrule
\multirow{2}{*}{Stack Overflow} & Train & 135.8 M & 342.5 K  & 1 & 183 & 194.2 K \\ \cmidrule(l){2-7}
    & Test & 16.6 M & 204.1 K & 1 & 43.2 K & 29 \\ \midrule
\multirow{2}{*}{CC-News} & Train & 661.6 K & 7.4 K & 1 & 16 & 24.4 K \\ \cmidrule(l){2-7}
    & Test & 45.3 K & 7.4 K & 1 & 2 & 32 \\ \bottomrule
\end{tabular}
\end{table}

To create a pre-training dataset with minimal privacy leakage, we start with the C4 dataset \citep{raffel2020exploring} and apply de-duplication techniques. First, we apply the \emph{approximate} duplicate detection method NearDup proposed by \citet{lee2022deduplicating} to filter out near-duplicates between C4 and the union of Stack Overflow and CC-News. To further remove potential overlap, we filter out all examples associated with \texttt{stackoverflow.com} or any URL in CC-News. This yields the \cmm dataset, which we use for pre-training. Statistics of example counts at each stage of the filtering pipeline are reported in \cref{cmm_stats}.

\begin{table}[htb]
\caption{Example counts from each stage of the pipeline to filter C4 and produce \cmm.}
\label{cmm_stats}
\begin{tabular}{@{}lcccc@{}}
\toprule
\textbf{Dataset} & \textbf{Split} & \textbf{\# Examples} & \textbf{\# Ex, de-duplicated} & \textbf{\# Ex, de-duplicated and URL-filtered} \\ \midrule
C4 & Train & 364.6 M & 345.6 M & 325.9 M \\ \bottomrule
\end{tabular}
\end{table}

\clearpage

\section{Configuring \elsfull}\label{appendix:configure_els}

\begin{figure}[ht]
    \centering
    \begin{subfigure}[b]{0.4\textwidth}
        \centering
        \includegraphics[width=\textwidth]{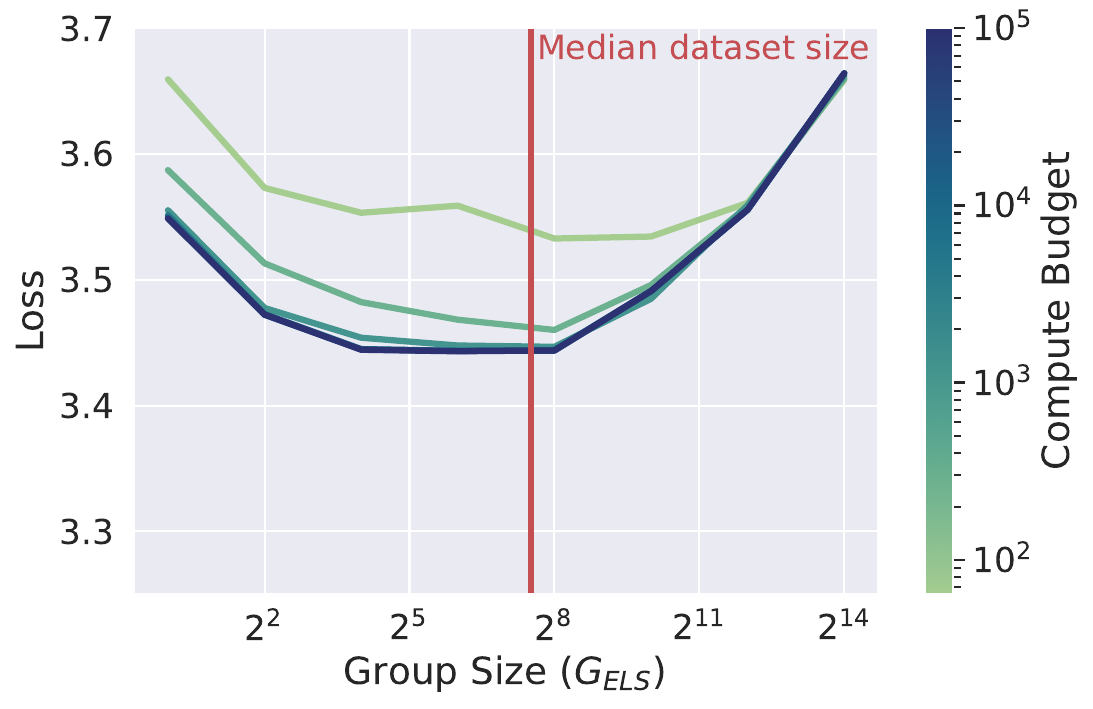}
        \caption{$\epsilon = 1$.}
        \label{fig:stackoverflow_els_eps_1}
    \end{subfigure}
    \hfill
    \begin{subfigure}[b]{0.4\textwidth}
        \centering
        \includegraphics[width=\textwidth]{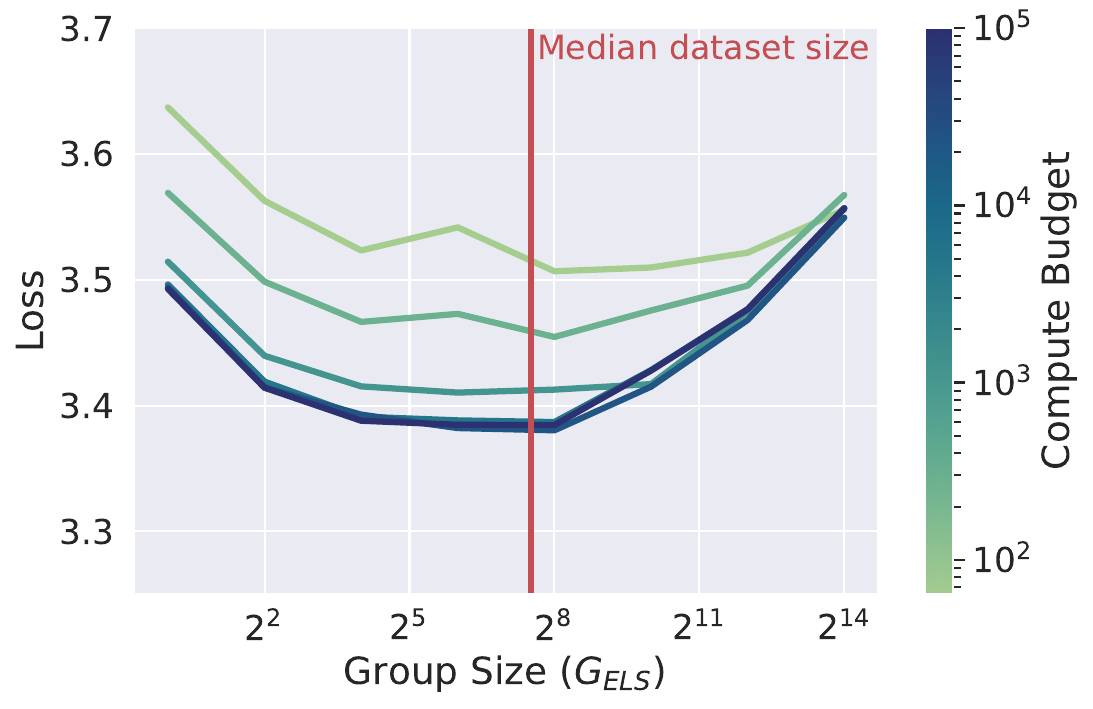}
        \caption{$\epsilon = 4$.}
        \label{fig:stackoverflow_els_eps_4}
    \end{subfigure}
    \hfill
    \begin{subfigure}[b]{0.4\textwidth}
        \centering
        \includegraphics[width=\textwidth]{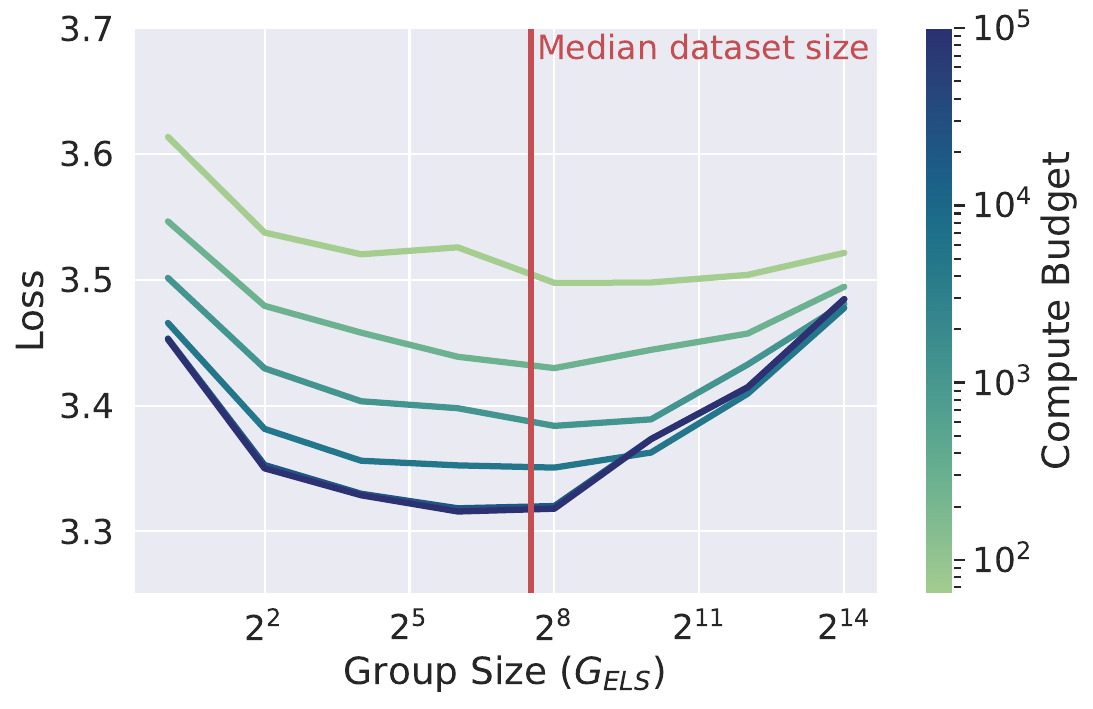}
        \caption{$\epsilon = 16$.}
        \label{fig:stackoverflow_els_eps_16}
    \end{subfigure}
    \hfill
    \begin{subfigure}[b]{0.4\textwidth}
        \centering
        \includegraphics[width=\textwidth]{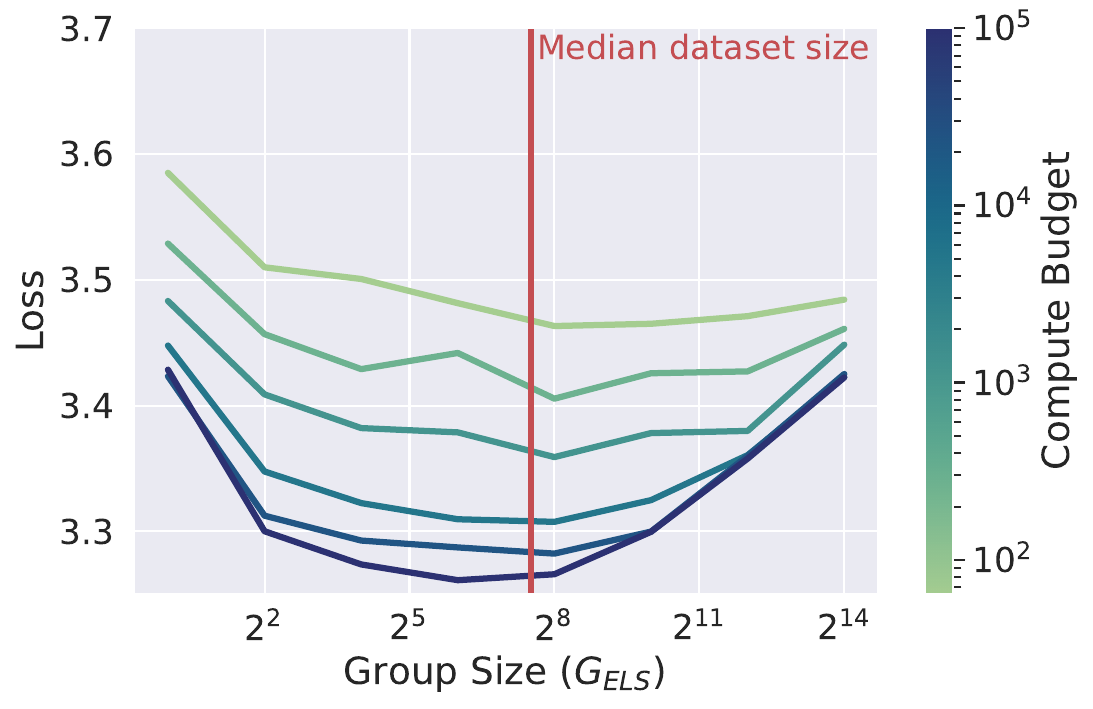}
        \caption{$\epsilon = 64$.}
        \label{fig:stackoverflow_els_eps_64}
    \end{subfigure}
    \caption{Loss of \els on Stack Overflow for varying $\epsilon$ and $G_\els$. The median user dataset size is plotted vertically.}
    \label{fig:stackoverflow_els_appendix}
    \vspace{-1cm}
\end{figure}

\begin{figure}[ht]
    \centering
    \begin{subfigure}[b]{0.4\textwidth}
        \centering
        \includegraphics[width=\textwidth]{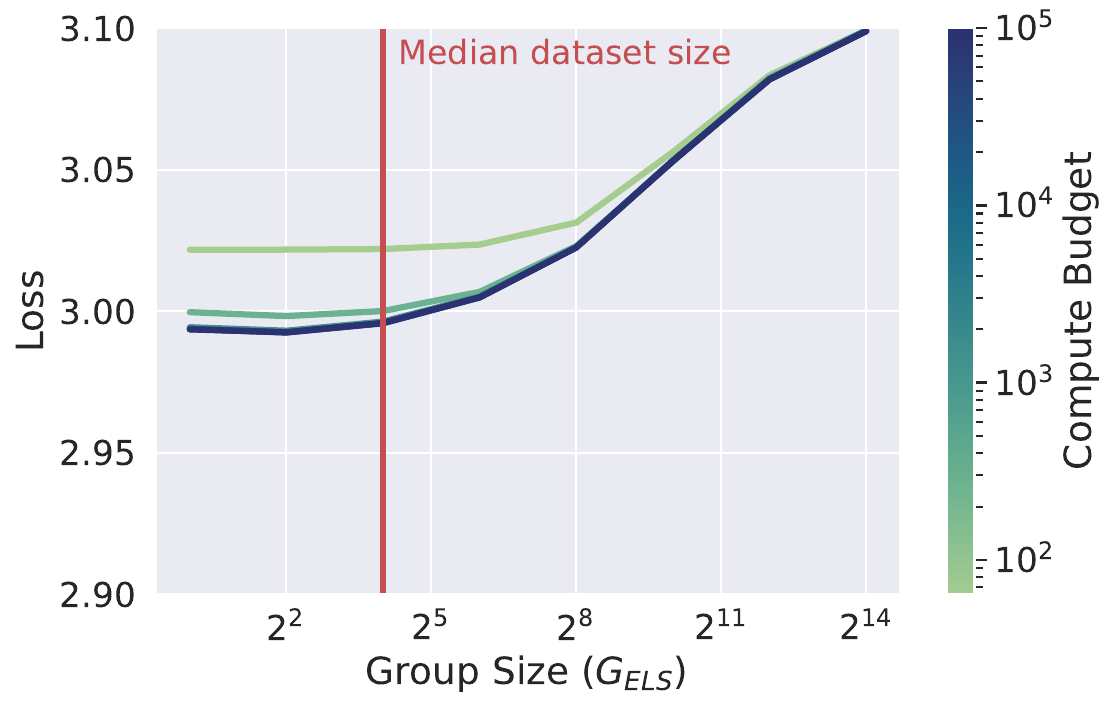}
        \caption{$\epsilon = 1$.}
        \label{fig:ccnews_els_eps_1}
    \end{subfigure}
    \hfill
    \begin{subfigure}[b]{0.4\textwidth}
        \centering
        \includegraphics[width=\textwidth]{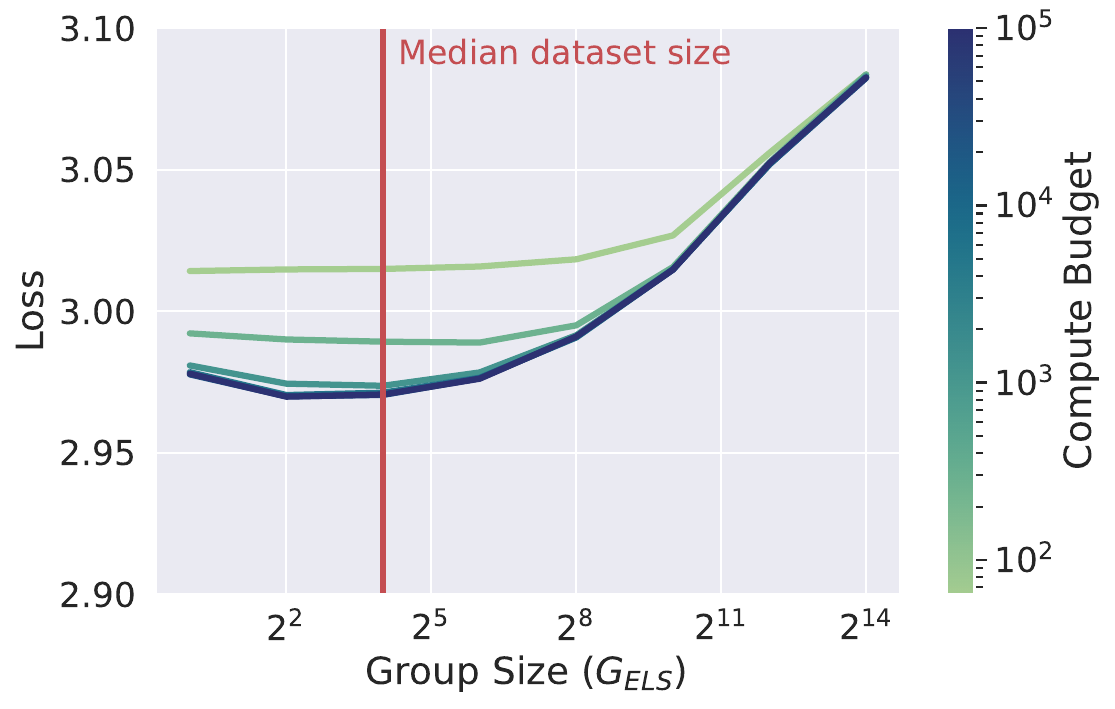}
        \caption{$\epsilon = 4$.}
        \label{fig:ccnews_els_eps_4}
    \end{subfigure}
    \hfill
    \begin{subfigure}[b]{0.4\textwidth}
        \centering
        \includegraphics[width=\textwidth]{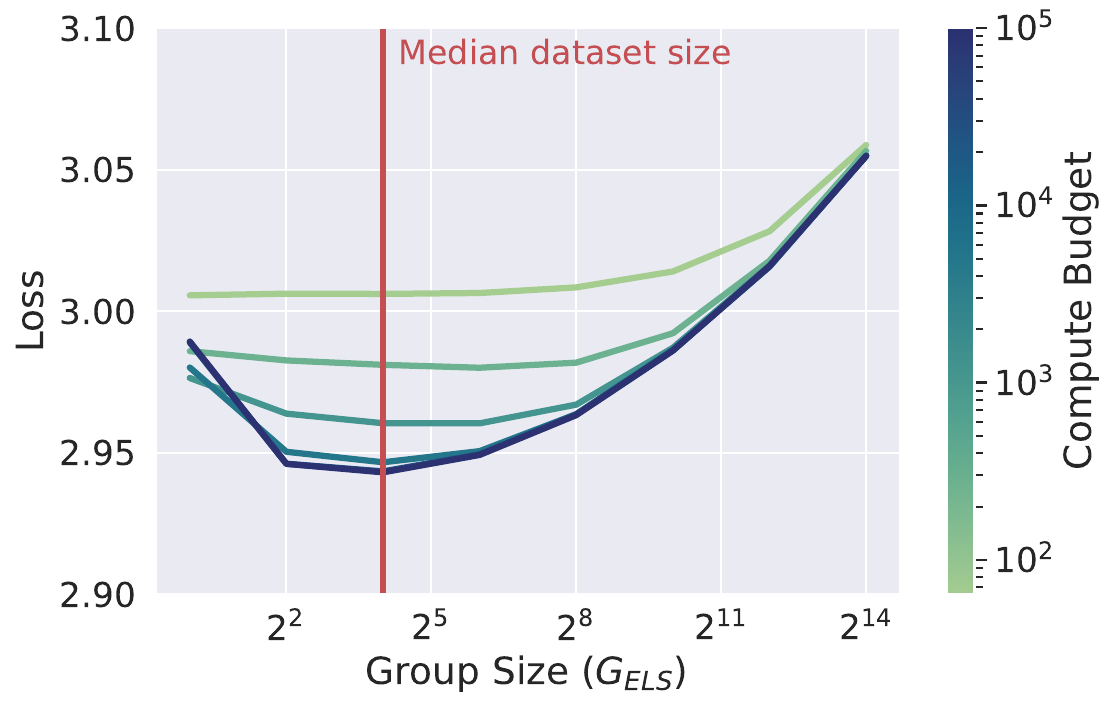}
        \caption{$\epsilon = 16$.}
        \label{fig:ccnews_els_eps_16}
    \end{subfigure}
    \hfill
    \begin{subfigure}[b]{0.4\textwidth}
        \centering
        \includegraphics[width=\textwidth]{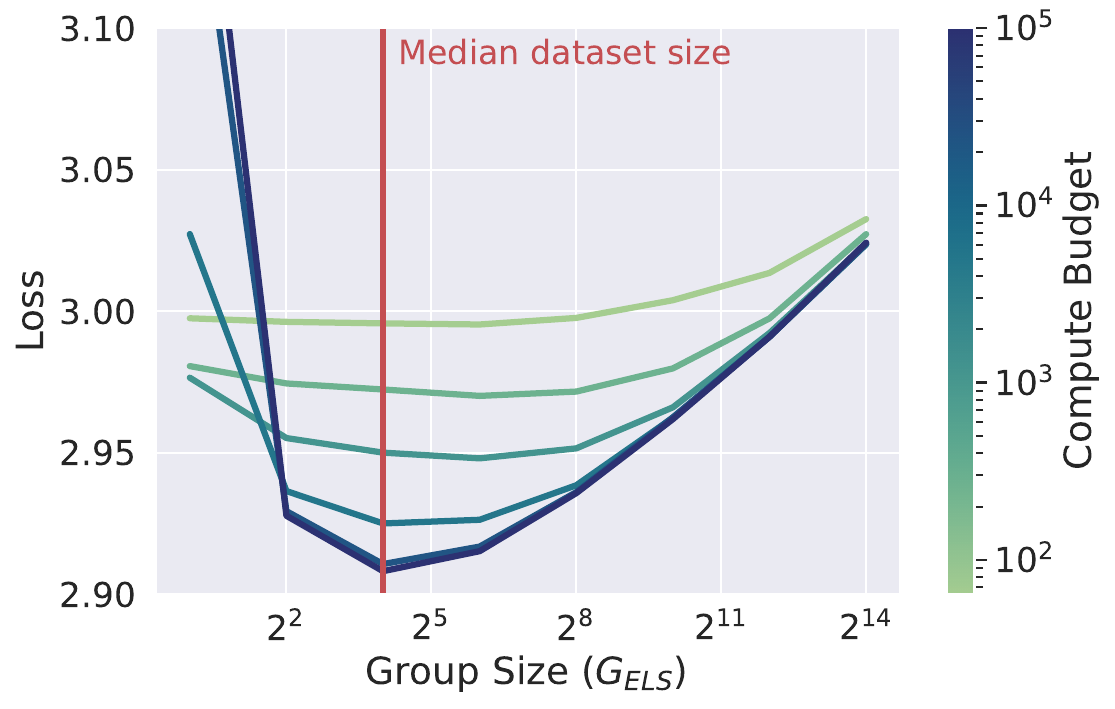}
        \caption{$\epsilon = 64$.}
        \label{fig:ccnews_els_eps_64}
    \end{subfigure}
    \caption{Loss of \els on CC-News for varying $\epsilon$ and $G_\els$. The median user dataset size is plotted vertically.}
    \label{fig:ccnews_els_appendix}
\end{figure}

In \cref{alg:dp-sgd-els}, $G_\els$ governs an important trade-off. Smaller values mean that \els trains on a small fraction of the dataset, while larger values means that users with more examples are potentially over-represented in sampling. To understand this, we fine-tune on both datasets using \els, for varying compute budget and group size. The results for Stack Overflow and CC-News are given in Figures \ref{fig:stackoverflow_els_appendix} and \ref{fig:ccnews_els_appendix}. While behavior for boundary values can be quite complicated and dependent on $\epsilon$, setting $G_\elst$ to the median dataset size works well throughout.

\clearpage

\section{Configuring \ulsfull}\label{appendix:configure_uls}

In \cref{sec:exp-results} we discussed the problem of how to set the group size parameter $G_\ulst$ for \uls. We expand on our discussion, and give a heuristic for selecting $G_\ulst$ for a given compute budget. Recall that in \cref{sec:understanding}, we showed that the variance of the additive noise in Algorithm \ref{alg:dp-sgd-uls}, which we denote $v_\ulst$, satisfies $v_\ulst \propto L_\ulst\sigma_\ulst$, where $\propto$ denotes proportionality, $L_\ulst$ is the maximum per-user gradient norm, and $\sigma_\ulst$ is the noise multiplier. We use $v_\ulst$ as a proxy for downstream performance.

As discussed in \cref{sec:understanding}, the quantity $L_\ulst$ is a function of $G_\ulst$, which we denote $L_\ulst(G_\ulst)$. To see this, note that each user-level gradient is an average over (at most) $G_\ulst$ example gradients. If, for example, every user has completely orthonormal example gradients, then $L_\ulst(G_\ulst) = 1/\sqrt{G_\ulst}$. The exact dependence of $L_\ulst$ on $G_\ulst$ is data-dependent, but can be estimated as a function of $G_\ulst$ by computing the norm of user-level gradients across the dataset. We do this as follows. We first sample some set $U$ of users. For each user $u \in U$, we randomly select a subset of its dataset of size (at most) $G_\ulst$. For each such user, we then compute
\[
\rho_u= \left\|\dfrac{1}{|D_u|}\sum_{z \in D_u} \nabla f(\theta, z)\right\|
\]
where $\theta$ is the pre-trained model. Let $\psi$ be some statistical estimator on sets of nonnegative real numbers. We can then approximate
\begin{equation}\label{eq:L_estimate}
L_\ulst(G_\ulst) \approx \psi(\{q_u\vert u \in U\}).
\end{equation}
As we discuss in \cref{sec:exp-results}, we let $\psi$ denote the median, rather than a maximum suggested by \eqref{eq:alpha_and_beta}. The noise multiplier $\sigma_\ulst$ is independent of $G_\ulst$, but depends on the sampling probability $q$ in Algorithm \ref{alg:dp-sgd-uls}. In settings with a fixed cohort size $M$, this means that $\sigma$ is a function of $M$, which we denote $\sigma_\ulst(M)$. We can compute this function via \eqref{eq:DP-SGD-ULS-accounting} using DP accounting libraries.

Fixing all other parameters of interest (including the desired privacy level $(\epsilon, \delta)$), the variance $v_\ulst$ of the noise added in \uls effectively satisfies the following:
\begin{equation}\label{eq:decomposable_variance}
v_\ulst (G_\ulst, M) \propto L_\ulst(G_\ulst)\sigma_\ulst(M).
\end{equation}

Now, say we have a desire compute budget $B$. To configure \uls, we must select a group size $G_\ulst$ and cohort size $M$ such that $G_\ulst M = B$. By \eqref{eq:decomposable_variance}, we would like to solve:
\begin{equation}\label{eq:decomposable_objective}
\min_{G_\ulst M = B} L_\ulst(G_\ulst)\sigma_\ulst(M)
\end{equation}

While we can compute $L(G_\ulst)$ and $\sigma_\ulst(M)$ at individual points, directly optimizing \eqref{eq:decomposable_objective} is challenging. Therefore, we consider a conceptually simpler problem: Suppose we are given some $G_\ulst$ and $M$, such that $G_\ulst M = B$. If we instead wanted to utilize a compute budget of $B' = 2B$, should we use the operating point $G_\ulst' = 2G_\ulst, M' = M$ or the operating point $G_\ulst' = G_\ulst, M' = 2M$? This can be answered by computing the following quantities:
\begin{equation}\label{eq:tau_objective}
\tau_G = \dfrac{L_\ulst(2G_\ulst)}{L_\ulst(G_\ulst)},~~~\tau_M = \dfrac{\sigma_\ulst(2M)}{\sigma_\ulst(M)}.
\end{equation}
Intuitively, these represent how much we shrink the objective in \eqref{eq:decomposable_objective} by doubling $G_\ulst$ or $M$, respectively. If $\tau_G < \tau_M$, then we should double $G_\ulst$, and otherwise we should double $M$. We formalize this iterative strategy in \cref{alg:configure_uls}, and refer to it as the ``Estimate-and-Double'' algorithm.

\begin{algorithm}[ht]
\caption{Configuring \ulsfull via ``Estimate-and-Double''}\label{alg:configure_uls}
\begin{algorithmic}
\State \textbf{Inputs:} Initial group size $G_\ulst^0$ and cohort size $M^0$, desired compute budget $B$.
\State{$G_\ulst \leftarrow G_\ulst^0, M \leftarrow M^0$.}
\State{Estimate $L_\ulst(G_\ulst)$ via \eqref{eq:L_estimate}, compute $\sigma(M)$ via \eqref{eq:DP-SGD-ULS-accounting}.}
\While{$G_\ulst M < B$}
  \State{Estimate $L_\ulst(G_\ulst)$ via \eqref{eq:L_estimate}, compute $\sigma(M)$ via \eqref{eq:DP-SGD-ULS-accounting}.}
  \State{$\tau_G \leftarrow \frac{L_\ulst(2G_\ulst)}{L_\ulst(G_\ulst)}$,~~~$\tau_M \leftarrow \frac{\sigma(2M)}{\sigma(M)}$.}
  \If{$\tau_G < \tau_M$}
    \State{$G_\ulst \leftarrow 2G_\ulst$}
  \Else
    \State{$M \leftarrow 2M$}
  \EndIf
\EndWhile
\end{algorithmic}
\end{algorithm}

\subsection{Validating Algorithm \ref{alg:configure_uls}}

To determine the efficacy of Algorithm \ref{alg:configure_uls}, we compute the loss of \uls when fine-tuning on Stack Overflow and CC-News, for a variety of group sizes $G_\ulst$ and cohort sizes $M$. We vary these over:
\begin{equation}\label{eq:uls_sweep}
G_\ulst \in \{2^0, 2^1, \dots, 2^8\},~~M \in \{2^5, 2^6,\dots, 2^{12}\},~~\epsilon \in \{1, 4, 16, 64\}.
\end{equation}

\begin{figure}[t]
    \centering
    \begin{subfigure}[b]{0.45\textwidth}
        \centering
        \includegraphics[width=\textwidth]{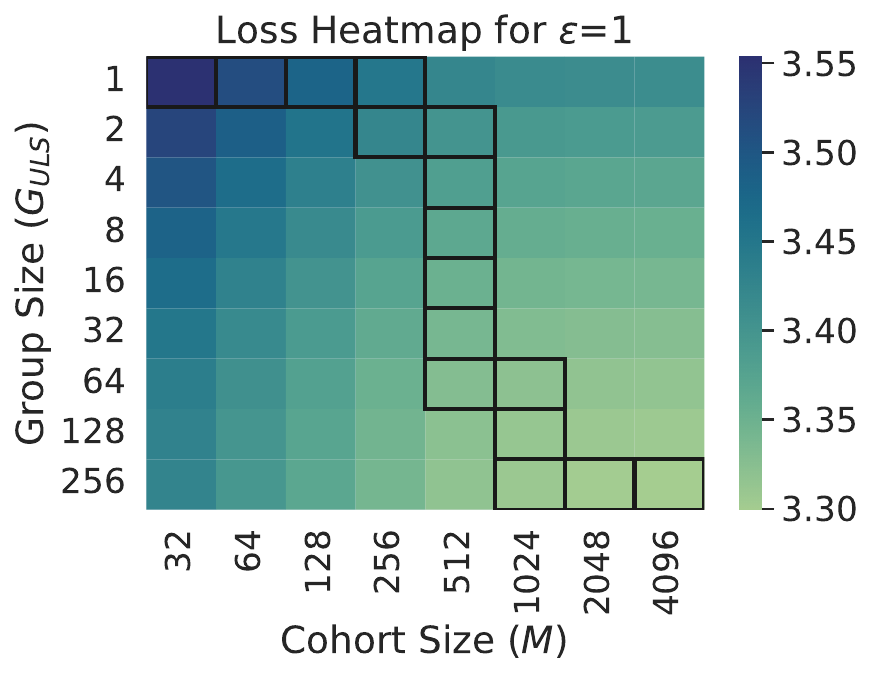}
        \label{fig:stackoverflow_heatmap_eps1}
    \end{subfigure}
    \hfill
    \begin{subfigure}[b]{0.45\textwidth}
        \centering
        \includegraphics[width=\textwidth]{plots/stackoverflow_heatmap_loss_epsilon4.pdf}
        \label{fig:stackoverflow_heatmap_eps4}
    \end{subfigure}
    \hfill
    \begin{subfigure}[b]{0.45\textwidth}
        \centering
        \includegraphics[width=\textwidth]{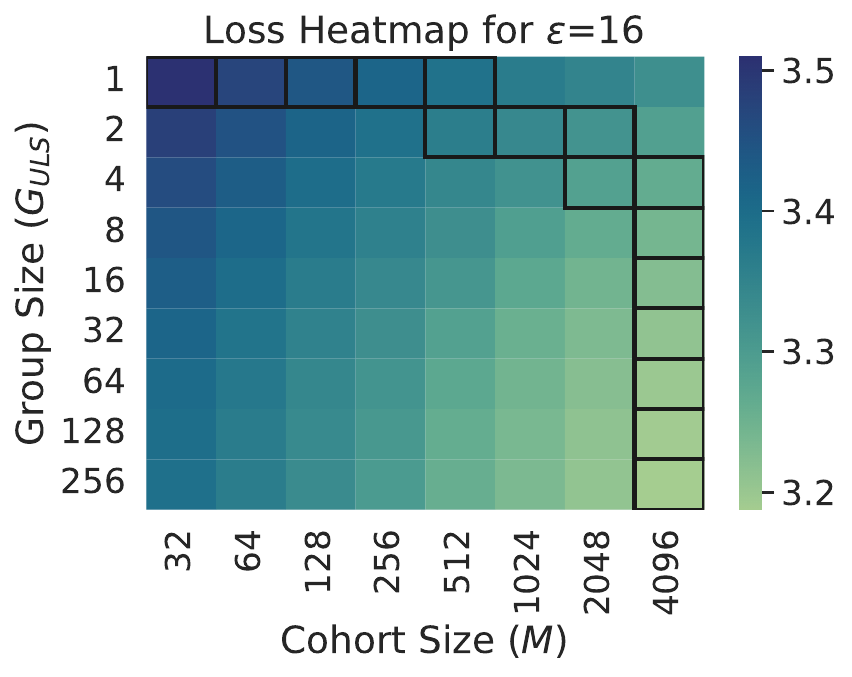}
        \label{fig:stackoverflow_heatmap_eps16}
    \end{subfigure}
    \hfill
    \begin{subfigure}[b]{0.45\textwidth}
        \centering
        \includegraphics[width=\textwidth]{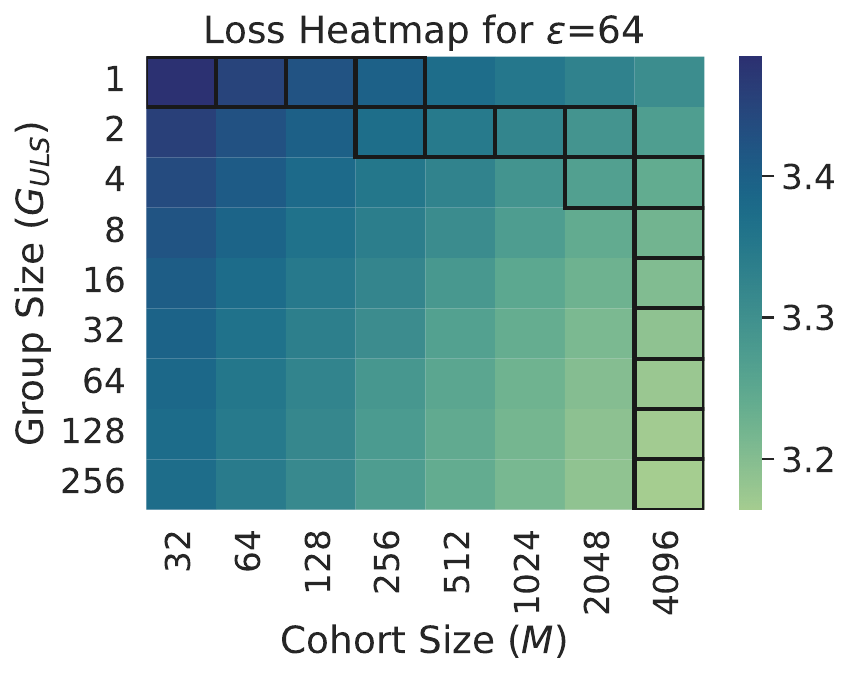}
        \label{fig:stackoverflow_heatmap_eps64}
    \end{subfigure}
    \caption{Loss heatmaps of \uls on Stack Overflow for varying group size $G_\uls$, cohort size $M$, and $\epsilon$. Optimal settings of $M$ and $G_\ulst$ for each compute budget are highlighted in black.}
    \label{fig:stackoverflow_uls_appendix}
\end{figure}

The results for Stack Overflow are given in \cref{fig:stackoverflow_uls_appendix}. We note that the anti-diagonals of the heatmaps represent a fixed compute budget. Using this information, we can now see how close the strategy in Algorithm \ref{alg:configure_uls} compares to the optimal setting of $G_\ulst$ and $M$ for a given compute budget (over the aforementioned powers of 2 we sweep over). Note that specifically, we initialize with $G_\ulst^0 = 1, M^0 = 32$.

To get a better sense of this, we compute, for compute budgets $B \in \{2^{5}, 2^{6}, \dots, 2^{20}\}$, the difference in loss between the optimal setting of $G_\ulst$ and $M$, and the following strategies:
\begin{itemize}
    \item \textbf{Greedy Local Oracle}: Given $G_\ulst, M$, this strategy has access to an oracle that can compute the fine-tuning loss when setting $G_\ulst' = 2G_\ulst, M' = M$, and when setting $G_\ulst' = G_\ulst, M' = 2M$. It then doubles whichever parameter results in a lower loss. Note that while this is not computationally tractable, it serves as a useful lower bound on the effectiveness of any local strategy.
    \item \textbf{Estimate-and-Double}: This is the strategy described by Algorithm \ref{alg:configure_uls}, starting with $G_\ulst^0 = 2^0, M^0 = 2^5$. We estimate $L_\ulst(G_\ulst)$ by sampling 128 users at random.
    \item \textbf{Random}: Given a compute budget $B$, this selects a random $G_\ulst, M$ from \eqref{eq:uls_sweep} such that $G_\ulst M = B$.
    \item \textbf{Max Cohort:} This picks $G_\ulst, M$ from \eqref{eq:uls_sweep} with a maximum value of $M$ such that $G_\ulst M = B$.
    \item \textbf{Max Group Size:} This picks $G_\ulst, M$ from \eqref{eq:uls_sweep} with a maximum value of $G_\ulst$ such that $G_\ulst M = B$.
\end{itemize}

\begin{figure}[htb]
    \centering
    \includegraphics[width=\linewidth]{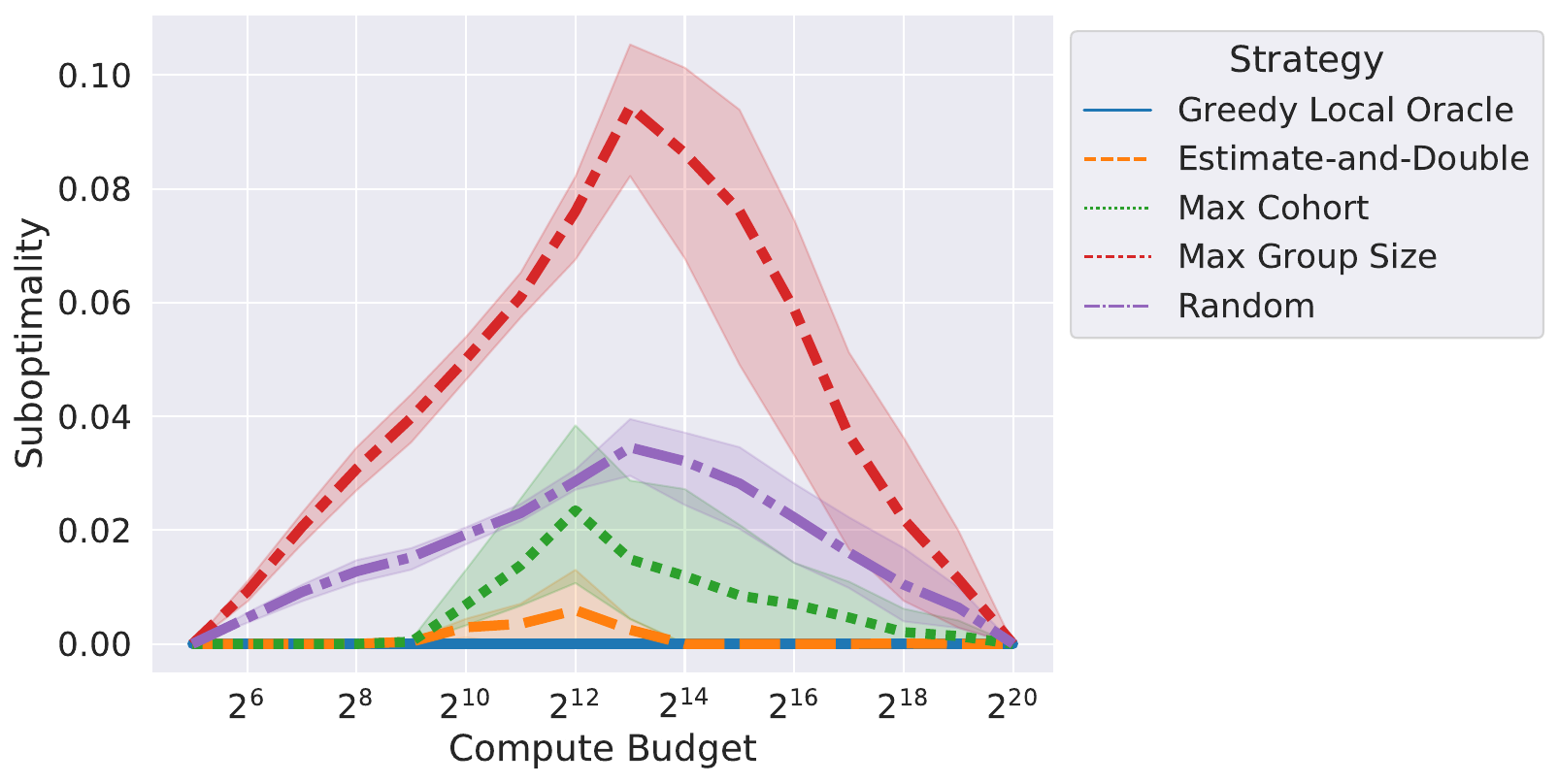}
    \caption{Sub-optimality (in terms of loss) for various strategies used to configure $G_\ulst, M$ in \ulsfull on Stack Overflow, for varying compute budgets. Results are averaged across $\epsilon \in\{1, 4, 16, 64\}$, and opaque areas represent the standard deviation.}
    \label{fig:stackoverflow_suboptimality}
\end{figure}

\begin{figure}[htb]
    \centering
    \includegraphics[width=\linewidth]{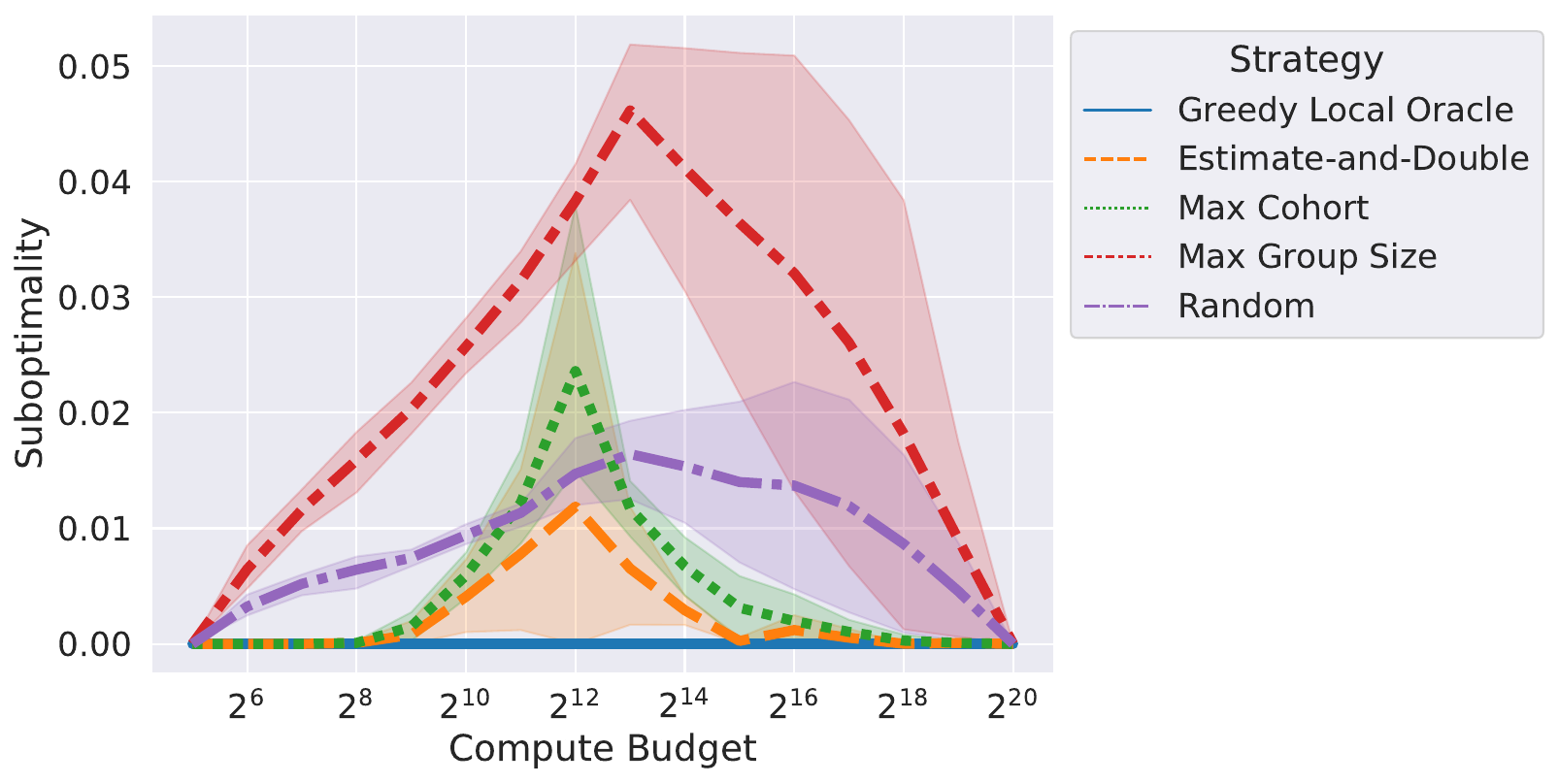}
    \caption{Sub-optimality (in terms of loss) for various strategies used to configure $G_\ulst, M$ in \ulsfull on CC-News, for varying compute budgets. Results are averaged across $\epsilon \in\{1, 4, 16, 64\}$, and opaque areas represent the standard deviation.}
    \label{fig:ccnews_suboptimality}
\end{figure}

The results are given in Figures \ref{fig:stackoverflow_suboptimality} and \ref{fig:ccnews_suboptimality}. Note that suboptimality here refers to the difference in loss between \ulsfull, when configured using one of the strategies above, versus when it is configured optimally. We see that \cref{alg:configure_uls} does well across compute budgets and $\epsilon$, for both datasets, and performs as well as the oracle strategy for most compute budgets.

\section{Additional Experimental Results}\label{appendix:additional_experiments}

\subsection{Compute-Loss Trade-offs}\label{appendix:compute_loss}

\begin{figure}[htb]
    \centering
    \includegraphics[width=\linewidth]{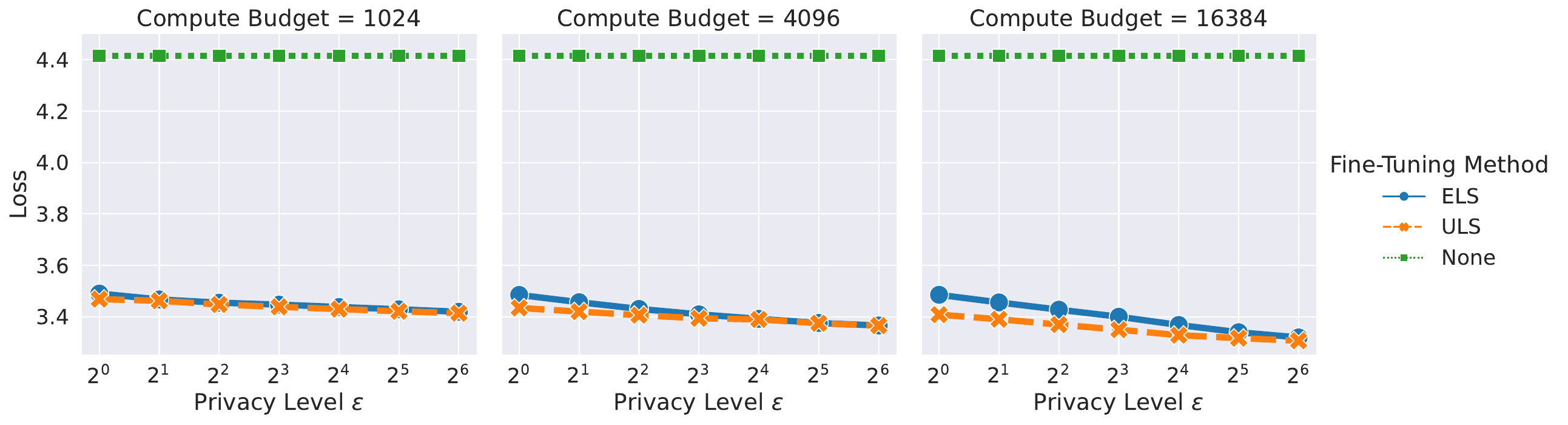}
    \caption{Compute-loss trade-offs on Stack Overflow, for varying privacy levels $\varepsilon$.}
    \label{fig:loss_budget_stackoverflow}
\end{figure}

\begin{figure}[htb]
    \centering
    \includegraphics[width=\linewidth]{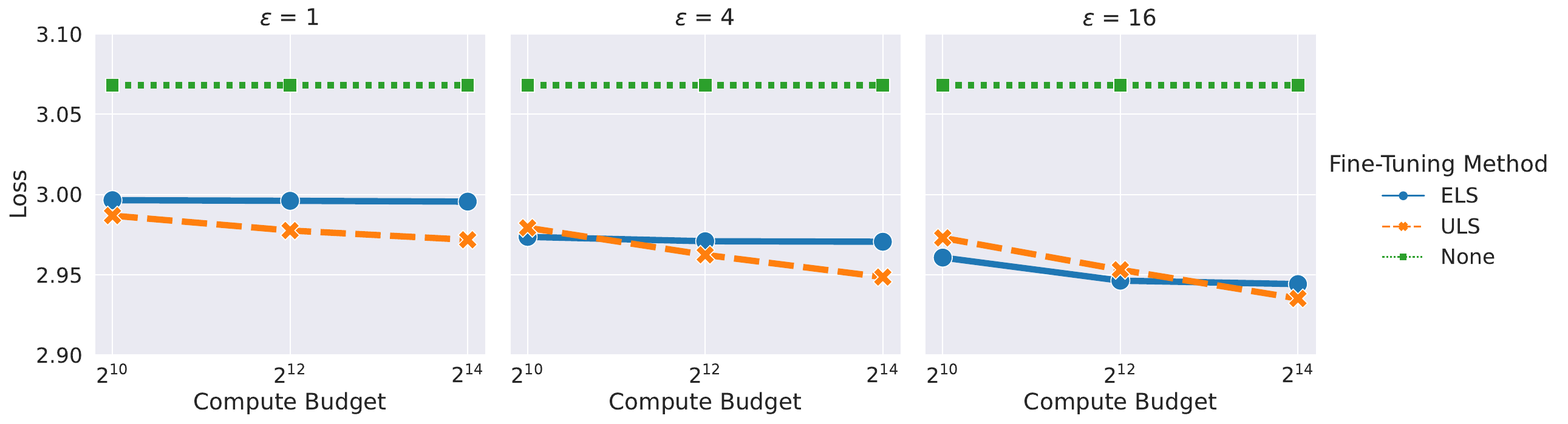}
    \caption{Compute-loss trade-offs on CC-News, for varying privacy levels $\varepsilon$.}
    \label{fig:loss_budget_ccnews}
\end{figure}

In Figures \ref{fig:loss_budget_stackoverflow} and \ref{fig:loss_budget_ccnews}, we present the same information as in Figures \ref{fig:loss_privacy_stackoverflow} and \ref{fig:loss_privacy_ccnews} in \cref{sec:exp-results}, but we instead consider the privacy-compute trade-offs for varying privacy levels $\varepsilon$. We find that \uls is more capable of improving its performance with increased compute budgets. In light of our analysis above, this makes intuitive sense: \els can only use increased compute budgets to reduce its noise multiplier $\sigma$, which has diminishing returns. However, \uls can allocate increased compute budgets to reduce its clip norm $C$ and its noise multiplier $\sigma$, trading them off as benefits saturate. We see the same effect in \cref{fig:compare_variance}.

\subsection{Personalizing Fine-Tuned Models}\label{appendix:personalization}

We take models trained via \els and \uls, and further personalize them to user data. In general, we are interested in whether the two algorithms exhibit different personalization behavior. A priori, this is plausible, as algorithms that operate at a user-level (including FedAvg~\citep{mcmahan2017communication}) can often exhibit improved personalization performance, even on LLM training tasks~\citep{charles2023towards}.

\begin{figure}[htb]
    \centering
    \includegraphics[width=\linewidth]{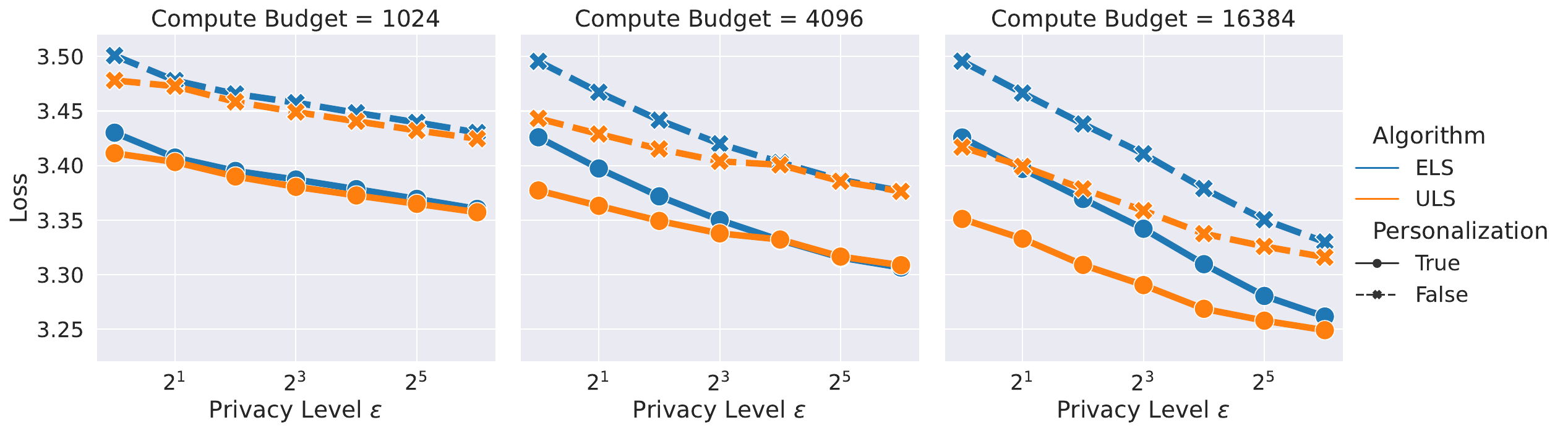}
    \caption{Privacy-loss trade-offs on Stack Overflow, for varying compute budgets, with and without personalization. We present the average loss across all test users on their held-out data.}
    \label{fig:loss_privacy_stackoverflow_personalization}
\end{figure}

To test this, we take our fine-tuned models, and further personalize them on each individual test user's dataset. We compare models fine-tuned via \els and \uls on the Stack Overflow dataset, and evaluate their personalization ability on the test users, comparing their performance with and without personalization. We do so by taking the Stack Overflow \els and \uls checkpoints, and further fine-tuning on individual test user datasets. We perform four local epochs of SGD, with a tuned learning rate, on half of each test users' examples. We then evaluate the personalized models on the reserved half of each test users' examples. We record the performance of each model, with and without personalization, on the reserved half of the test users' examples. The results are in Figure \ref{fig:loss_privacy_stackoverflow_personalization}.

We see that for both algorithms, personalization seems to incur a uniform reduction in loss. However, the gap between using and not using personalization seems to be roughly the same for both model checkpoints. Personalization does not seem to change the fundamental shape of the trade-off curves. In particular, \uls with personalization seems to outperform or match \els for the same privacy levels and compute budgets as without personalization.

\end{document}

%% file: macros.tex
\newcommand{\cmm}{C4-{}-\xspace}

\newcommand{\CommentStyle}[1]{\textcolor{blue}{\textit{#1}}}
\newcommand{\eat}[1]{}

\newtheorem{conj}{Conjecture}
\newtheorem{lem}{Lemma}
\newtheorem{thm}{Theorem}

\newtheorem{obs}{Observation}

\renewcommand{\epsilon}{\varepsilon}

\DeclareMathOperator{\clip}{clip}
\DeclareMathOperator{\bin}{Binom}
\DeclareMathOperator{\median}{median}
\DeclareMathOperator{\bern}{Bern}
\DeclareMathOperator{\variance}{var}
\DeclareMathOperator{\sym}{sym}

\newcommand{\calA}{\mathcal{A}}

\newcommand{\calM}{\mathcal{M}}
\newcommand{\calN}{\mathcal{N}}

\newcommand{\calU}{\mathcal{U}}

\newcommand{\boldx}{\ensuremath{\boldsymbol{x}}}

\newcommand{\R}{\mathbb{R}}

\newcommand{\E}{\mathbb{E}}

\newcommand{\ltwo}[1]{\left\|#1\right\|_2}
\newcommand{\elsfull}{\texttt{DP-SGD-ELS}\xspace}
\newcommand{\els}{\texttt{ELS}\xspace}
\newcommand{\elst}{\text{ELS}}
\newcommand{\ulsfull}{\texttt{DP-SGD-ULS}\xspace}
\newcommand{\uls}{\texttt{ULS}\xspace}
\newcommand{\ulst}{\text{ULS}}

\newcommand{\mypar}[1]{\smallskip
	\noindent{\textbf{{#1}.}}}